\documentclass{article}

\usepackage{microtype}
\usepackage{graphicx}
\usepackage{subcaption}
\usepackage{booktabs} 

\usepackage{hyperref}

\usepackage[preprint]{icml2026}

\usepackage{amsmath}
\usepackage{amssymb}
\usepackage{mathtools}
\usepackage{amsthm}
\usepackage{algorithmic}
\usepackage{algorithm}
\usepackage{authblk}

\usepackage[capitalize,noabbrev]{cleveref}

\theoremstyle{plain}
\newtheorem{theorem}{Theorem}[section]
\newtheorem{proposition}[theorem]{Proposition}

\theoremstyle{definition}
\newtheorem{definition}[theorem]{Definition}
\newtheorem{assumption}[theorem]{Assumption}
\theoremstyle{remark}
\newtheorem{remark}[theorem]{Remark}

\usepackage{amsmath,amsfonts,bm}

\def\eqref#1{equation~\ref{#1}}

\def\1{\bm{1}}

\DeclareMathAlphabet{\mathsfit}{\encodingdefault}{\sfdefault}{m}{sl}
\SetMathAlphabet{\mathsfit}{bold}{\encodingdefault}{\sfdefault}{bx}{n}

\def\gI{{\mathcal{I}}}

\newcommand{\KL}{D_{\mathrm{KL}}}

\usepackage[textsize=tiny]{todonotes}
\usepackage{tikz}

\definecolor{bricksTitleBlue}{RGB}{238,247,253}

\title{\bf BRICKS-WM: Building Reusability via Interface Composition Kinetics for Structured World Models}
\date{}

\newcommand\nnfootnote[1]{%
  \begin{NoHyper}
  \renewcommand\thefootnote{}\footnote{#1}%
  \addtocounter{footnote}{-1}%
  \end{NoHyper}
}

\author[1,2]{\bf Shaowei Zhang}
\author[1,2]{\bf Jiahan Cao}
\author[3]{\bf Xunlan Zhou}
\author[1,2]{\bf Shenghua Wan}
\author[1,2]{\bf De-Chuan Zhan$^\dag$}

\affil[1]{National Key Laboratory for Novel Software Technology, Nanjing University, China}
\affil[2]{School of Artificial Intelligence, Nanjing University, China}
\affil[3]{School of Intelligence Science and Technology, Nanjing University, China}

\affil[ ]{\small\texttt{\{zhangsw, caojh, wansh\}@lamda.nju.edu.cn, wyattzhouxl@smail.nju.edu.cn, zhandc@nju.edu.cn }}

\makeatletter
\icml@noticeprintedtrue
\hypersetup{
  pdfauthor={Shaowei Zhang, Jiahan Cao, Xunlan Zhou, Shenghua Wan, De-Chuan Zhan},
  pdftitle={BRICKS-WM: Building Reusability via Interface Composition Kinetics for Structured World Models},
  pdfsubject={},
  pdfkeywords={}
}
\fancypagestyle{bricksfirstpage}{%
  \fancyhf{}%
  \renewcommand{\headrulewidth}{0pt}%
  \fancyfoot[C]{\thepage}%
}
\newcommand{\bricksTitleBlock}{%
  \begingroup
  \noindent
  \begin{tikzpicture}
    \node[
      fill=bricksTitleBlue,
      rounded corners=8pt,
      inner sep=0pt,
      text width=\textwidth
    ] {%
      \begin{minipage}{\textwidth}
        \vspace{0.18in}
        \begin{center}
          \begin{minipage}{0.94\linewidth}
            \centering
            {\LARGE\bfseries\@title\par}
            \vspace{0.12in}
            {\normalsize\@author\par}
          \end{minipage}
        \end{center}
        \vspace{0.02in}
        \begin{center}
          \begin{minipage}{0.94\linewidth}
            Model-based Reinforcement Learning (MBRL) has achieved remarkable success in continuous control by leveraging latent world models. However, prevailing approaches typically rely on monolithic latent dynamics, entangling environment dynamics into a coupled process. This coupling severely limits reusability: altering the agent necessitates retraining the entire world from scratch, even if the environment remains constant.
To address this, we introduce \textbf{BRICKS-WM} (\emph{\textbf{B}uilding \textbf{R}eusability via \textbf{I}nterface \textbf{C}omposition \textbf{K}inetics for \textbf{S}tructured \textbf{W}orld \textbf{M}odels}), a framework for the modular assembly of structured world models.
Driven by the insight that the physical world is composed of independent entities, we posit that global dynamics can be modeled as a composition of distinct dynamical modules interacting via latent interfaces.
As a minimal instantiation, we factorize the latent state space into an actuated Agent module and an external Background module, bridged by a learned latent interface.
Unlike prior object-centric methods that prioritize visual segmentation, BRICKS-WM enforces a functional separation in transition dynamics, ensuring that background dynamics remains agnostic to the agent's dynamics.
Empirically, BRICKS-WM achieves control performance comparable to strong monolithic baselines when trained from scratch, and enables the reuse of frozen background dynamics across agents.

          \end{minipage}
        \end{center}
        \vspace{0.08in}
      \end{minipage}%
    };
  \end{tikzpicture}%
  \par
  \endgroup
}
\makeatother

\begin{document}

\onecolumn
\thispagestyle{bricksfirstpage}
\bricksTitleBlock
\nnfootnote{$^\dag$ Corresponding author.}

\section{Introduction}
\label{sec:intro}

Model-based Reinforcement Learning (MBRL) has achieved remarkable success by learning latent world models that facilitate planning in low-dimensional spaces ~\cite{planet, dreamerv3, muzero, tdmpc2}. By compressing high-dimensional sensory data into compact latent states, these agents can effectively imagine future outcomes to master complex control tasks. 
However, despite their performance, a recurring limitation remains: most latent world models operate as \emph{monolithic} black boxes whose latent state entangles distinct functional components, such as controllable entities and environmental physics. This coupling creates a fundamental inefficiency: when the scenario changes, the learned dynamics of the world cannot be reused. Consequently, researchers are forced to retrain the entire world model from scratch for new task scenarios, even if the underlying dynamical subsystems remain identical.

In this paper, we adopt a different perspective: rather than treating the world model as a monolithic predictor, we study \emph{modular assembly of structured world models}. We posit that a scalable world model should not be a black box, but rather a composition of distinct, reusable dynamical modules that can be assembled to construct new environments. Under this framework, global dynamics emerge from the interaction of independent subsystems via standardized interfaces, allowing specifically learned dynamics to be recomposed across tasks and scenarios, analogous to the assembly of physical bricks. In this view, the key technical challenge is not merely to learn factorized representations or dynamics~\cite{sear, dear, tia, denoisedmdp, iso_dream}, but to learn dynamics modules whose interfaces support recombination without breaking predictive consistency. This modular objective aligns with broader trends in compositional representation learning and object-centric modeling~\cite{slot_attention,iodine,monet,genesis,space,savi,slotssm} and with modularity for transfer in control~\cite{modular_policy,smp,metamorph,cross_embodiment_wm_2025}. However, existing object-centric world models for control tasks typically remain functionally coupled: even when the latent state is factorized, the transition model is optimized as a single joint predictor, which makes it difficult to freeze and reuse a subset of the dynamics in a new scenario. The closest modular mechanism approach is COMET~\cite{comet}, which promotes decompositions, but does not directly address cross-task reuse by freezing a subset of dynamics and re-learning only the necessary components under an explicit protocol constraint.

As a foundational step towards this vision, we introduce \textbf{BRICKS-WM} (\emph{\textbf{B}uilding \textbf{R}eusability via \textbf{I}nterface \textbf{C}omposition \textbf{K}inetics for \textbf{S}tructured \textbf{W}orld \textbf{M}odels}). While the principle of modularity is general, this paper operationalizes it through a minimal yet meaningful instantiation: factorizing the world into an \textit{Actuated Agent} module and an \textit{External Background} module. 
Unlike prior object-centric approaches that focus primarily on visual segmentation \cite{slot_attention, savi, monet, sold}, BRICKS-WM enforces a functional separation in transition \textit{dynamics}.  We propose a latent interface for the interaction between the agent and the background. 
By constraining the background dynamics to be conditional solely on this interface and structurally independent of the agent's specific state, we transform the background dynamics into a reusable module. This allows a background model trained with one agent to be frozen and reused by a new agent in related settings, effectively turning the transfer problem into a protocol matching problem.
While sharing conceptual similarities with modular transfer in policy learning~\cite{modular_policy,smp}, our work is distinguished by its focus on \emph{modular reuse at the dynamics level}. Specifically, we propose reusing components of the world model rather than transferring policies.

Our approach bridges the gap between structured representations and high-performance control. While recent works have explored object-centric world models~\cite{dyno, sold, fiocwm}, they typically employ coupled transition functions, which restrict the ability to independently reuse dynamics modules across different tasks. Our specific contributions are as follows:

\begin{itemize}
    \item \textbf{Framework for Modular Dynamics Assembly}: We propose BRICKS-WM, a framework designed to assemble structured world models from reusable dynamical modules. As a foundational instantiation, we factorize global dynamics into Agent and Background dynamics mediated by a latent interface. Distinct from prior methods, this functional separation ensures background dynamics remain agnostic to the agent's dynamics to facilitate modular reuse.
    
    \item  \textbf{SOTA-Level Control with Object-Centric Models}: We demonstrate that object-centric structure need not compromise control efficacy. When trained from scratch, BRICKS-WM achieves performance on par with the monolithic state-of-the-art baselines, DreamerV3~\cite{dreamerv3} and TD-MPC2~\cite{tdmpc2}. To the best of our knowledge, this is the \emph{\textbf{first}} object-centric world model approach to match them on dynamic \emph{\textbf{locomotion}} tasks solely from raw pixels, without ground-truth state supervision.
    \item  \textbf{First Demonstration of Modular Dynamics Reuse}: We introduce a mechanism for transfer by freezing the background module and aligning the new agent via the latent interface. This enables the reuse of environmental dynamics without retraining the whole world model, yielding competitive control performance especially in related target settings.
\end{itemize}

\section{Preliminaries}
\label{sec:preliminaries}

\paragraph{Problem formulation.}
We model the environment as a Partially Observable Markov Decision Process (POMDP) defined by the tuple
$\mathcal{M}=(\mathcal{S},\mathcal{A},\mathcal{O},\mathcal{T},\Omega,\mathcal{R},\gamma)$,
where $\Omega$ denotes the observation, or emission, kernel. At time $t$, the environment is in an unobserved ground-truth state
$s_t^*\in\mathcal{S}$ and emits an observation $o_t\sim\Omega(\cdot\mid s_t^*)$. Given the history
$\mathcal{H}_t\triangleq(o_{1:t},a_{1:t-1})$, the agent executes an action $a_t\sim\pi(\cdot\mid\mathcal{H}_t)$,
receives a reward $r_t\sim\mathcal{R}(\cdot\mid s_t^*,a_t)$, and causes the state to evolve as
$s_{t+1}^*\sim\mathcal{T}(\cdot\mid s_t^*,a_t)$. The next observation is then generated as
$o_{t+1}\sim\Omega(\cdot\mid s_{t+1}^*)$. We define a trajectory as
$\tau=\{(o_t,a_t,r_t)\}_{t=1}^{T}$. The standard control objective is to learn a policy
$\pi(a_t\mid\mathcal{H}_t)$ that maximizes the expected return
$\mathbb{E}_{\pi,\mathcal{M}}\big[\sum_{k=0}^{\infty}\gamma^k r_{t+k}\big]$.

\paragraph{Compositional Reuse Setting.}
Unlike standard monolithic modeling, we approach the problem through \emph{modular composition}. We posit that global dynamics arise from the interaction of distinct, reusable dynamics. While this principle is general, this work focuses on a minimal instantiation: separating the \emph{actuated agent} from the \emph{external background}. We define a transfer scenario involving a source task $\mathcal{M}_{src}$ and a target task $\mathcal{M}_{tgt}$. Crucially, these tasks share a similar background dynamics module (governing external objects and terrain) but employ distinct agent modules or tasks.
Our goal is to enable reuse of the background dynamics model learned from $\mathcal{M}_{src}$, re-assembling it with an agent dynamics module trained from scratch in $\mathcal{M}_{tgt}$ under a standardized interaction protocol, thereby avoiding the need to re-learn the whole environment from scratch.

\paragraph{Recurrent State-Space Model.}
To handle high-dimensional observations in POMDP, we leverage the Recurrent State-Space Model (RSSM)~\cite{planet}. The RSSM maintains a deterministic recurrent state $h_t$ and a stochastic latent state $s_t$. The dynamics contains a recurrent model $h_t = f_{\theta}(h_{t-1}, s_{t-1}, a_{t-1})$ and a stochastic transition model $p_{\theta}(s_t\mid h_t)$. For notational brevity, we omit explicit references to $h_t$ in the subsequent sections. While effective, standard RSSMs learn a monolithic state representation $s_t$. In this work, we adopt the RSSM backbone but modify the internal factorization to enable the modular reuse described above.

\section{Related Work}
\label{sec:related_work}

\paragraph{Latent World Models for Control.} Model-based reinforcement learning (MBRL) has achieved significant success by learning latent dynamics models that facilitate planning in low-dimensional spaces. Dreamer series~\cite{dreamerv1, dreamerv2, dreamerv3} establishes a strong baseline by learning deterministic and stochastic components of the environment dynamics. MuZero~\cite{muzero} achieves superhuman performance via planning with a learned model on Atari and board games such as Go, SimPLe~\cite{atari100k} demonstrates sample-efficient model-based RL on Atari 100k, and TD-MPC~\cite{tdmpc, tdmpc2} learns task-oriented latent dynamics for short-horizon MPC in continuous control. Recent advances have also improved model expressivity via Transformer-based world models \cite{iris,storm} and large-scale generative interactive environments \cite{genie}, as well as diffusion-based future modeling beyond step-by-step rollout~\cite{diffusion_world_model} and diffusion-based world modeling that emphasizes high-fidelity visual prediction~\cite{diamond}.
However, despite their diversity, these approaches typically learn a monolithic latent state that entangles the agent's dynamics with the environment. This coupling necessitates retraining the entire world model when the agent changes, whereas BRICKS-WM proposes a modular factorization to enable the direct reuse of environmental dynamics.

\paragraph{Object-Centric Representation and Structured Dynamics.} To improve generalization and reasoning, a growing body of work focuses on decomposing scenes into distinct entities using object-centric inductive biases. Foundational methods such as Slot Attention~\cite{slot_attention}, IODINE~\cite{iodine}, MONet~\cite{monet}, SPACE~\cite{space}, and GENESIS~\cite{genesis} demonstrate how to unsupervisedly segment static scenes into constituent slots. In the temporal domain, models like SAVi~\cite{savi} and PSB~\cite{psb_2024} extend these representations to video. Related lines of work model physical interactions with relational or interaction-network inductive biases~\cite{interaction_networks, nri, op3, cswm, rims}. Separately, COMET~\cite{comet} studies modular world models via independent mechanisms.
Recent methods have successfully integrated these structured representations into control, as seen in Dyn-O~\cite{dyno}, SOLD~\cite{sold}, and FIOCWM~\cite{fiocwm}.
 However, while these approaches leverage factorization to improve physical reasoning, they typically learn dynamics that remain functionally coupled, preventing independent module reuse. COMET~\cite{comet} studies modular mechanism discovery, but does not explicitly evaluate freeze-and-transplant reuse of a learned dynamics submodule under a transfer protocol. In contrast, our framework prioritizes dynamics composition, enabling the construction of world models from reusable modules, instantiated here via an Agent-Background factorization linked by a standardized interface.

\paragraph{Modular Transfer and Cross-Embodiment Generalization.}
Cross-embodiment transfer has been extensively explored from the perspective of policy learning.
Techniques such as modular policy decompositions~\cite{modular_policy} and Shared Modular Policies ~\cite{smp} enable transfer by separating task-specific goals from robot-specific actuation. More recently, large-scale pretraining has sought to learn universal controllers~\cite{morphology_agnostic,metamorph} or cross-embodiment world models for specific domains like manipulation~\cite{cross_embodiment_wm_2025}.
Uniquely, BRICKS-WM targets modular world-model reuse. Rather than transferring a policy or learning a universal model from massive datasets, we propose a mechanism to freeze and transplant a pretrained background dynamics module. By relearning only agent-side components under a standardized protocol, we enable dynamics reuse in target tasks without the need for large-scale multi-robot data.

\section{Method}
\label{sec:method}
\begin{figure*}[htp!]
\centerline{\includegraphics[width=\linewidth]{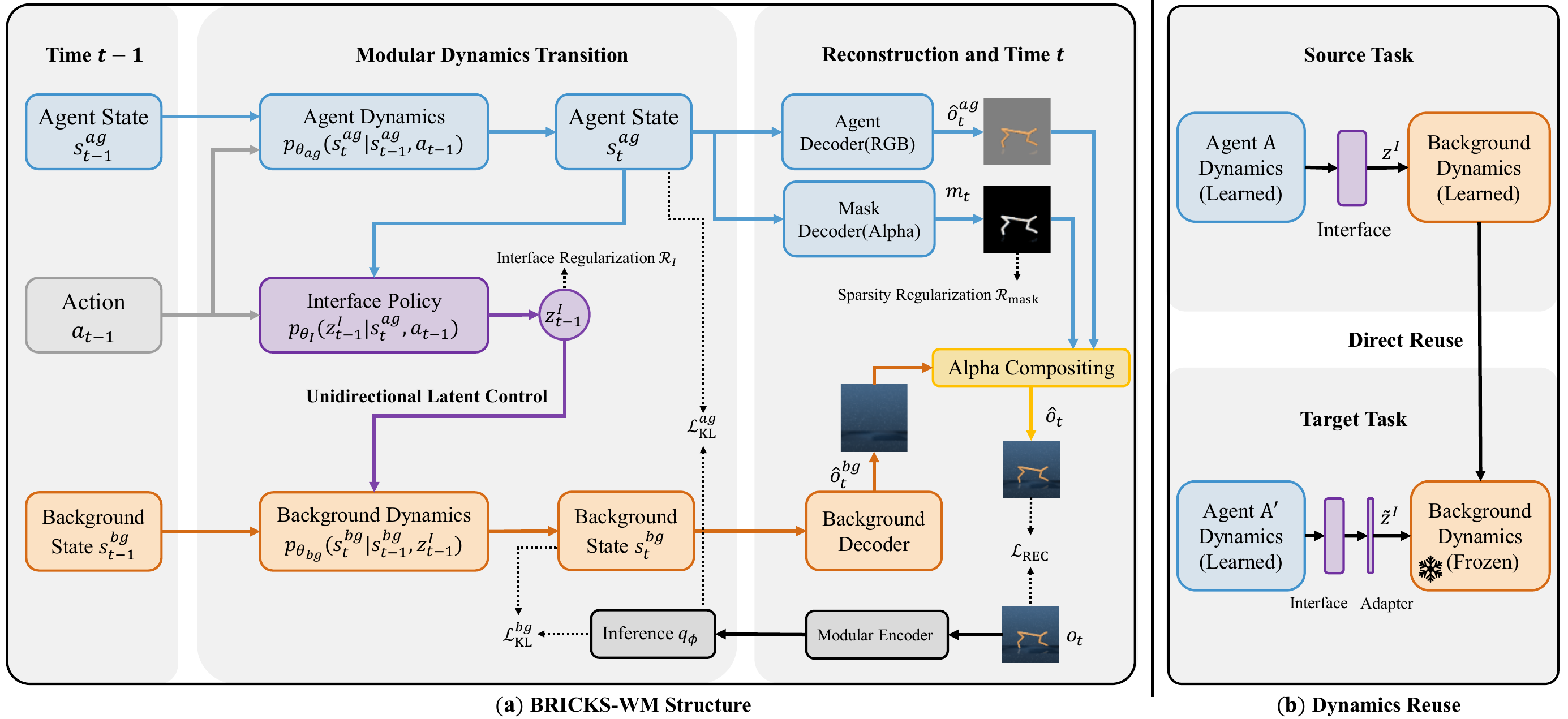}}
\caption{\textbf{Framework of BRICKS-WM.} (a) The overall architecture decomposes global dynamics into Agent and Background modules mediated by a latent interface. (b) The dynamics reuse logic, illustrating how a frozen background module is transferred to a new agent in a target task via an interface adapter.}
\label{fig:framework}
\end{figure*}

We propose BRICKS-WM (\emph{\textbf{B}uilding \textbf{R}eusability via \textbf{I}nterface \textbf{C}omposition \textbf{K}inetics for \textbf{S}tructured \textbf{W}orld \textbf{M}odels}), a framework designed to enable \emph{modular reuse} in model-based reinforcement learning.
Unlike monolithic world models that learn entangled dynamics end-to-end, BRICKS-WM fundamentally rethinks the world model as a composition of distinct, reusable dynamics interacting via a standardized latent protocol.
As a foundational instantiation of this framework, we focus on a minimal decomposition: factorizing the environment into an actuated \emph{Agent} module and an external \emph{Background} module. These modules are bridged by a learned, standardized latent interface. This functional factorization allows the Background dynamics to be encapsulated as a reusable transition module, which can be transferred across related agents and tasks without re-learning the environment from scratch. In the following, we detail the entity representation~(Sec.~\ref{sec:structural_disentaglement}), interface-composed dynamics~(Sec.~\ref{sec:composable_dynamics}), training objectives~(Sec.~\ref{sec:training_obj}) and the reuse mechanism~(Sec.~\ref{sec:module_reuse}). The framework is shown in Figure~\ref{fig:framework}.

\subsection{Structural Disentanglement of Entity Representations}
\label{sec:structural_disentaglement}
To enable modular reuse, we need to ensure that the latent space does not implicitly entangle agent and environment features. Instead of a monolithic representation, we explicitly decouple the global state into independent latent subspaces for the Agent and the Background, interacting solely through a standardized bottleneck. Crucially, this separation serves a functional purpose: it isolates the background dynamics into a reusable module. This distinguishes our approach from prior factorized methods~\cite{iso_dream, denoisedmdp}, as we transfer the background dynamics module to related target tasks via a shared protocol.

\begin{definition}[Structured latent state with interface]
At time $t$, the latent state is factorized into
$s_t \triangleq (s^{ag}_{t}, s^{bg}_{t}),$
where $s^{ag}_{t}\in\mathbb{R}^{d_{ag}}$ encodes agent features and
$s^{bg}_{t}\in\mathbb{R}^{d_{bg}}$ encodes the background information.
An \emph{interface} latent $z^{\mathcal{I}}_{t}\in\mathbb{R}^{d_{\mathcal{I}}}$ mediates the agent's control over the background.
\end{definition}

Given an observation $o_t$, we infer two entity slots via a modular encoder. We leverage Slot Attention~\cite{slot_attention} (implementation details can refer to Appendix~\ref{app:modular_encoder_datail}) to enforce object binding, yielding a factorized posterior:
\begin{equation}
q_{\phi}(s^{ag}_{t}, s^{bg}_{t} \mid o_t)
= q_{\phi_{ag}}(s^{ag}_{t} \mid o_t) \cdot q_{\phi_{bg}}(s^{bg}_{t} \mid o_t).
\label{eq:factorized_posterior}
\end{equation}
This factorization acts as an inductive bias, where the background representation remains conditionally independent of the agent's identity during posterior inference.

\begin{remark}
    This factorization adopts the assumption of statistical independence to enforce strict disentanglement during perception, ensuring that the background representation remains agnostic to the agent's identity to facilitate direct modular reuse in target domains.
\end{remark}
\paragraph{Asymmetric composite decoder.}
To reconstruct the observation $o_t$, we employ an asymmetric decoding strategy that reflects the physical composition of the scene.
The agent state $s^{ag}_t$ is decoded into an RGBA component (color image $\hat{o}^{ag}_t$ and an alpha mask $m_t$), while the background state $s^{bg}_t$ is decoded strictly into a RGB background $\hat{o}^{bg}_t$:
\begin{align}
    (\hat{o}^{ag}_t, m_t) &= D_{\theta_{ag}}(s^{ag}_t), \quad \text{where } m_t \in [0, 1]^{H \times W}, 
    \hat{o}^{bg}_t = D_{\theta_{bg}}(s^{bg}_t).
\end{align}
The final predicted image $\hat{o}_t$ is formed via alpha compositing:
\begin{equation}
    \hat{o}_t = m_t \odot \hat{o}^{ag}_t + (1 - m_t) \odot \hat{o}^{bg}_t.
    \label{eq:alpha_blend}
\end{equation}
We use reconstruction to learn an explicit composable world model; in richer visual scenes, this choice may waste capacity on task-irrelevant appearance and can be replaced by non-reconstruction objectives.

Because Slot Attention is permutation-equivariant, to encourage the semantic binding of slots, we introduce a regularization term. We add an L1 penalty term for $m_t$ as a sparsity regularizer to minimize the locations that are identified as regions of agents:
\begin{equation}
\mathcal{R}_{\texttt{mask}} = \sum_{t=1}^{T} \|m_t\|_1
\label{eq:mask_reg}
\end{equation}

Crucially, this perceptual disentanglement is a prerequisite for modular reuse. If the observation encoder entangles agent-specific features with background appearance, the reused background model fails to generalize to new agents. By explicitly routing information into distinct slots and decoding asymmetrically, we ensure that the interface and dynamics factorization remain well-posed for transfer.

\subsection{Composable Dynamics via an Action-Conditioned Interface}
\label{sec:composable_dynamics}

Standard latent dynamics models (e.g., RSSM~\cite{planet}) learn a joint transition $p(s_{t}\mid s_{t-1}, a_{t-1})$, inextricably coupling the agent and the background.
We instead propose an \emph{interface-composed} factorization that transforms the background dynamics into a distinct, controllable module.

\paragraph{Generative dynamics factorization.}
To enable modular transfer, the background dynamics must be decoupled from the agent's specific dynamics. Previous methods like Iso-Dream~\cite{iso_dream} assume that the two dynamics are completely independent. Considering the interactions between dynamics, we propose the following Unidirectional Latent Control assumption to achieve a balance:

\begin{assumption}[Unidirectional Latent Control]
\label{ass:uni_interface}
The interaction between modules is mediated solely by the latent interface variable $z^{\mathcal{I}}_t$.
Crucially, we assume the background state evolution is \emph{action-agnostic} conditioned on this interface:
$s^{bg}_{t}$ depends on $(s^{bg}_{t-1}, z^{\mathcal{I}}_{t-1})$, but is conditionally independent of the agent's specific action $a_{t-1}$ or state $s^{ag}_{t-1}$.
\end{assumption}

Based on Assumption~\ref{ass:uni_interface}, the joint transition does not entangle the agent and background directly. Instead, it factorizes into three conditionally independent modules chained by the interface:
\begin{align}
 p_{\theta}\big(s^{ag}_{t}, s^{bg}_{t}, z^{\mathcal{I}}_{t-1} \mid s^{ag}_{t-1}, s^{bg}_{t-1}, a_{t-1}\big)
& =
\underbrace{p_{\theta_{ag}}(s^{ag}_{t} \mid s^{ag}_{t-1}, a_{t-1})}_{\text{agent dynamics}}
\cdot
\underbrace{p_{\theta_{\mathcal{I}}}(z^{\mathcal{I}}_{t-1} \mid s^{ag}_{t}, a_{t-1})}_{\text{interface policy}}
\nonumber \\
& \quad \cdot
\underbrace{p_{\theta_{bg}}(s^{bg}_{t} \mid s^{bg}_{t-1}, z^{\mathcal{I}}_{t-1})}_{\text{background dynamics}}.
\label{eq:dynamics_factorization}
\end{align}

Each component plays a distinct role in our modular architecture:
\begin{itemize}
    \item \textbf{Agent Dynamics} $\theta_{ag}$: models the kinematic transition of the agent. This module operates in isolation, independent of the background state.
    \item \textbf{Interface Policy} $\theta_{\mathcal{I}}$: is the key bridging module. It infers the abstract interaction signal $z^{\mathcal{I}}_{t-1}$ (e.g., forces, ego-motion) given the agent's action $a_{t-1}$ and its resultant state $s^{ag}_{t}$. Conditioning on the \textit{next} state $s^{ag}_{t}$ allows the policy to generate signals consistent with the realized movement.
    \item \textbf{Background Dynamics} $\theta_{bg}$: simulates the transition of the global scene driven by $z^{\mathcal{I}}_t$. Since this module does not directly observe $a_t$, it remains agnostic to the agent's dynamics, supporting reuse across different agents.
\end{itemize}
\begin{remark}[Implicit Feedback vs. Explicit Coupling]
One might argue that physical interactions are bidirectional (e.g., ground reaction forces). However, explicitly modeling the \emph{background $\rightarrow$ agent} feedback creates a tight coupling that hinders modularity.
Instead, we rely on the \emph{implicit representation capacity} of the agent's state as a trade-off:
\begin{itemize}
    \item \textbf{Implicit Encoding:} Any physical feedback from the environment is naturally captured in the posterior $q(s^{ag}_t \mid o_{\le t})$ inferred from observations. The agent dynamics $p_{\theta_{ag}}$ learns to model these effects implicitly without requiring a separate channel from $s^{bg}_t$.
    \item \textbf{Explicit Control:} The background dynamics remains controllable via $z^{\mathcal{I}}_t$, allowing the agent to influence the scene without breaking the independence required for reuse.
\end{itemize}
This design effectively trades physical completeness for representational disentanglement, and may be limited in settings with rich bidirectional interactions.
\end{remark}

\begin{definition}[Reusable dynamics module via protocol matching]
A dynamics module $p_{\theta_{bg}}(s^{bg}_{t}\mid s^{bg}_{t-1}, z^{\mathcal{I}}_{t-1})$ is \emph{reusable} across agents if, for a new agent, there exists an interface generator producing $z^{\mathcal{I}}_{t-1}$ such that the induced interface process matches the protocol expected by the module (i.e., aligning with the distribution learned during source domain training).
\end{definition}

\begin{proposition}[Sufficient condition for background-module reuse]
\label{prop:reuse}
Consider two agents $A$ (source) and $A'$ (target) operating with the same frozen background module $p_{\theta_{bg}}(s^{bg}_{t}\mid s^{bg}_{t-1}, z^{\mathcal{I}}_{t-1})$.
If for all $t$, the conditional interface distributions match under comparable histories:
\begin{equation}
p^{A}(z^{\mathcal{I}}_{t-1}\mid \mathcal{H}_{t-1})
= p^{A'}(z^{\mathcal{I}}_{t-1}\mid \mathcal{H}_{t-1}),
\label{eq:interface_match}
\end{equation}
then the induced background transition distributions also match:
\begin{equation}
p^{A}(s^{bg}_{t}\mid s^{bg}_{t-1}, \mathcal{H}_{t-1})
= p^{A'}(s^{bg}_{t}\mid s^{bg}_{t-1}, \mathcal{H}_{t-1}).
\label{eq:bg_match}
\end{equation}
\end{proposition}

Proposition~\ref{prop:reuse} formalizes our key insight: \emph{cross-task reuse is fundamentally a protocol-matching problem}. The sufficient condition for reuse is not merely matching a canonical prior, but matching the interface process expected by the frozen background module. In particular, if the target agent induces the same conditional interface distribution as the source agent under comparable histories, then the frozen background transition induces the same conditional background dynamics. The shared canonical prior $\mathcal{P}_{prior}$, such as $\mathcal{N}(0,I)$, is therefore used as a practical regularization target rather than as a sufficient condition by itself. Specifically, our KL regularization term $\mathcal{R}_{\gI}$ in Sec.~\ref{sec:training_obj} provides a capacity constraint and canonicalization bias: it discourages the interface from carrying arbitrary agent-specific information and empirically facilitates protocol alignment. The residual adapter introduced in Sec.~\ref{sec:module_reuse} further compensates for remaining protocol shifts between source and target agents.

\subsection{Training Objective}
\label{sec:training_obj}

We optimize a variational evidence lower bound (ELBO) on the action-conditioned likelihood of observation sequences. The objective is derived from the interplay between our structured generative model and a factorized inference network.

\paragraph{Generative model.}
We define the joint distribution over observations $o_{1:T}$ and structured latent variables, with Agent states $s^{ag}_{1:T}$, Background states $s^{bg}_{1:T}$, and Interface latents $z^{\mathcal{I}}_{0:T-1}$. Leveraging the compositional factorization from  Eq.~\ref{eq:dynamics_factorization}, the generative process is given by:
\begin{equation} \label{eq:bricks_joint_long}
\begin{aligned}
p_{\theta}\big(o_{1:T}, s^{ag}_{1:T}, s^{bg}_{1:T}, z^{\mathcal{I}}_{0:T-1}\mid a_{1:T}\big)
&= \prod_{t=1}^{T} \bigg[ p_{\theta_o}(o_t\mid s^{ag}_t,s^{bg}_t) \cdot p_{\theta_{ag}}(s^{ag}_{t}\mid s^{ag}_{t-1},a_{t-1})
\\
&\quad \cdot p_{\theta_{\mathcal{I}}}(z^{\mathcal{I}}_{t-1}\mid s^{ag}_{t},a_{t-1}) \cdot p_{\theta_{bg}}(s^{bg}_{t}\mid s^{bg}_{t-1},z^{\mathcal{I}}_{t-1}) \bigg].
\end{aligned}
\end{equation}

\paragraph{Inference Model.}
To approximate the posterior, we employ a variational encoder $q_{\phi}$ that factorizes consistently with the generative structure:
\begin{equation} \label{eq:bricks_posterior}
\begin{aligned}
q_{\phi}\big(s^{ag}_{1:T}, s^{bg}_{1:T}, z^{\mathcal{I}}_{0:T-1}\mid o_{1:T}, a_{1:T}\big)
&= \prod_{t=1}^{T} \bigg[ q_{\phi_{ag}}(s^{ag}_{t}\mid s^{ag}_{t-1},a_{t-1},o_{t}) \\
&\quad \cdot q_{\phi_{\mathcal{I}}}(z^{\mathcal{I}}_{t-1}\mid s^{ag}_{t},a_{t-1})
\cdot q_{\phi_{bg}}(s^{bg}_{t}\mid s^{bg}_{t-1},z^{\mathcal{I}}_{t-1},o_{t}) \bigg].
\end{aligned}
\end{equation}

Crucially, the interface posterior $q_{\phi_{\mathcal{I}}}$ is inferred solely from the agent's transitions. This reflects our design that interaction events are implicit in the agent's kinematics, preventing the interface from peeking at the background state directly.

\paragraph{Variational Lower Bound (ELBO).}
Maximizing the log-likelihood is bounded by minimizing the negative ELBO, which decomposes into reconstruction and dynamics consistency terms:
\begin{equation}
\log p_{\theta}(o_{1:T}\mid a_{1:T})
\ge
\mathcal{L}_{\texttt{ELBO}}
\triangleq
\sum_{t=1}^{T}\mathbb{E}_{q_{\phi}}\!\left[\log p_{\theta_o}(o_t\mid s^{ag}_t,s^{bg}_t)\right]
-\mathcal{R}_{\textsc{kl}} .
\label{eq:bricks_elbo_main}
\end{equation}

where $\mathcal{R}_{\textsc{KL}}$ sums the KL divergences for the Agent and Background modules (see Appendix A for full derivation).
\begin{equation}
\begin{aligned}
\mathcal{R}_{\textsc{kl}}
\triangleq
\sum_{t=1}^{T}\mathbb{E}_{q}\Big[
&\KL\!\big(q_{\phi_{ag}}(s^{ag}_{t}\!\mid s^{ag}_{t-1},a_{t-1},o_t)
\,\|\,p_{\theta_{ag}}(s^{ag}_{t}\!\mid s^{ag}_{t-1},a_{t-1})\big)\\
&+
\KL\!\big(q_{\phi_{bg}}(s^{bg}_{t}\!\mid s^{bg}_{t-1},z^{\mathcal{I}}_{t-1},o_t)
\,\|\,p_{\theta_{bg}}(s^{bg}_{t}\!\mid s^{bg}_{t-1},z^{\mathcal{I}}_{t-1})\big)
\Big].
\label{eq:bricks_kl_decompose}
\end{aligned}
\end{equation}

\paragraph{Interface standardization as a protocol regularizer.}
While standard ELBO maximization inherently minimizes the KL divergence for latent variables, our specific parameterization causes the interface KL term to vanish (see Appendix~\ref{app:elbo} for derivation). Consequently, to prevent the interface from carrying arbitrary information, we explicitly re-introduce a regularization term:
\begin{equation}
\mathcal{R}_{\gI}
=
\sum_{t=1}^{T}\mathbb{E}_{q}\Big[
\KL\!\big(q_{\phi_{\mathcal{I}}}(z^{\mathcal{I}}_{t-1}\!\mid s^{ag}_{t},a_{t-1})\,\|\,\mathcal{N}(0,I)\big)
\Big],
\label{eq:proto_kl_planet}
\end{equation}
By limiting the capacity of the interface channel, this term functions as an \emph{Information Bottleneck}. Since $z^{\mathcal{I}}$ is the \emph{sole} channel modulating the background, the capacity constraint forces the interface to encode only the most salient, domain-invariant variations (e.g., global displacement, force magnitude) necessary to minimize background reconstruction error. Conversely, agent-specific details (e.g., limb articulation, local texture), which provide no gain for background prediction, are actively filtered out to conserve the information budget. Thus, $\mathcal{R}_{\gI}$ acts not merely as a prior constraint, but as a \emph{semantic filter}, driving the emergence of a standardized, transferable interaction protocol.

\paragraph{Overall training loss.}
The total objective combines the ELBO with the structural constraints derived from our sparsity principle (Eq.~\ref{eq:mask_reg}) and protocol standardization (Eq.~\ref{eq:proto_kl_planet}).

\begin{equation}
\mathcal{L} = - \mathcal{L}_{\text{ELBO}} + \beta_{\texttt{mask}}  \mathcal{R}_{\texttt{mask}} + \beta_{\gI} \mathcal{R}_{\gI}.
\end{equation}

\subsection{Module Reuse via Compositional Transfer}
\label{sec:module_reuse}

As shown in Figure~\ref{fig:framework}(b), to transfer the encapsulated environmental dynamics to a new task, we perform a compositional transfer. We instantiate a target BRICKS-WM by using the pretrained background subsystem from a source model and re-learning only the agent-side components under the standardized interface.

\paragraph{Initialization and Parameter Inheritance.}
Concretely, we initialize the target model by selectively importing parameters $\theta^{src}_{\texttt{share}}$ corresponding to the \textbf{domain-invariant} subsystems: the perceptual encoder (including slot decomposition), the background dynamics and the background decoder. Conversely, agent-specific components $\theta^{tgt}_{\texttt{new}}$ (agent dynamics, interface policy, reward/continuation heads) are re-initialized from scratch.

\paragraph{Freezing policy.}
For transfer, we freeze the core transition mechanism of the background dynamics (e.g., the GRU recurrent core in RSSM) to preserve the learned background dynamics. While the core transition module is locked, we allow the background observation heads mild adaptability to re-calibrate to the new agent's visual occlusion patterns.

\begin{figure*}[ht!]
\centerline{\includegraphics[width=\linewidth]{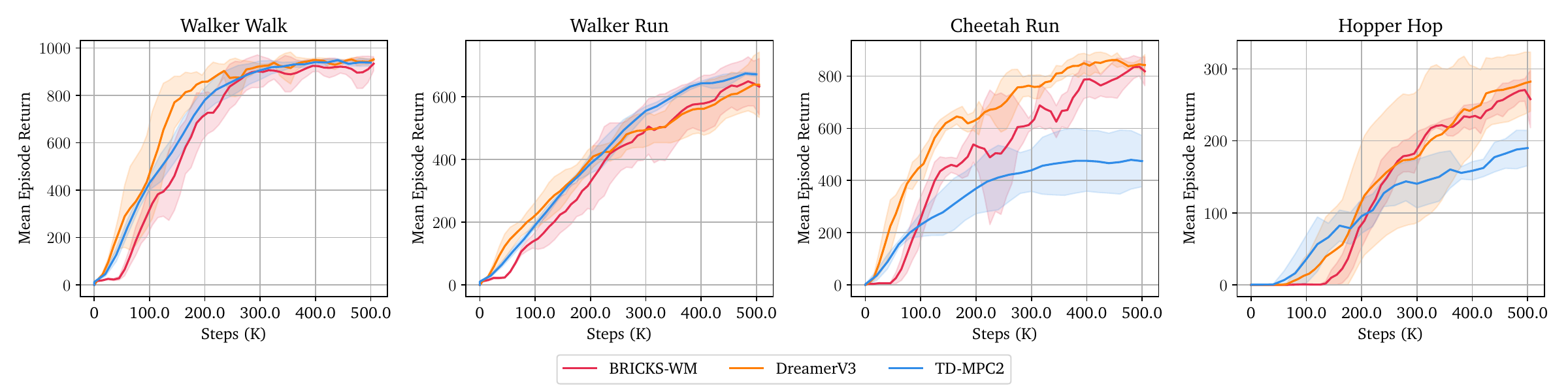}}
\caption{\textbf{Performance on DMC trained from scratch.}}
\label{fig:scratch_train_performance}
\end{figure*}

\begin{figure*}[htp!]
\centerline{\includegraphics[width=\linewidth]{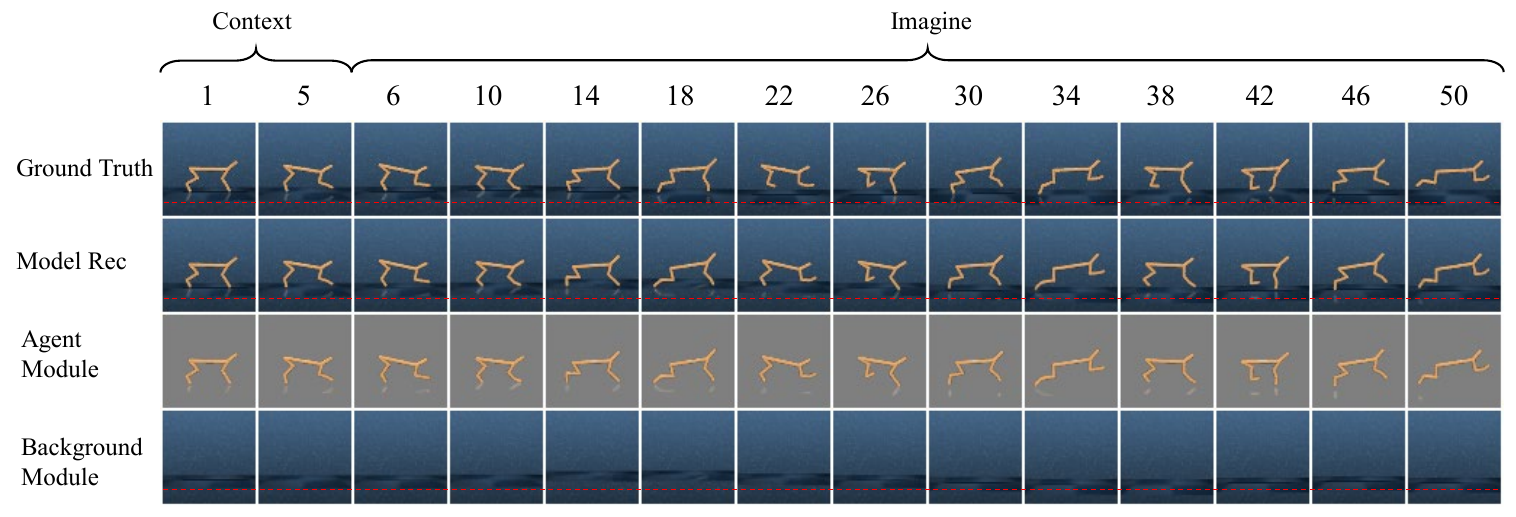}}
\caption{
\textbf{Latent modulation analysis on \emph{Cheetah Run}.}
    The model receives the first 5 frames as context and performs open-loop imagination from timestep $t=6$. During imagination, we modulate one dimension of the interface latent $z^{\mathcal{I}}$ with a sinusoidal signal.
    \textbf{Rows 1--4} show the ground truth, full reconstruction, agent component, and background component, respectively.
    The \textbf{red dashed lines} highlight the vertical floor oscillation induced by the modulation. Notably, unlike the ground truth (\textbf{Row 1}), both the full reconstruction (\textbf{Row 2}) and background (\textbf{Row 4}) display clear vertical floor oscillations.
}

\label{fig:visualization_cheetah_sine_wave}
\end{figure*}

\paragraph{Identity-residual interface adapter.}
Although the interface is standardized via $\mathcal{R}_{\gI}$, minor distribution shifts may persist between different agent morphologies. To facilitate precise protocol alignment without disrupting the pretrained dynamics, we insert a two-layer lightweight residual adapter $A_{\psi}$ before the frozen background module:
\begin{equation}
\tilde{z}^{\mathcal{I}}_t = A_{\psi}(z^{\mathcal{I}}_t) = z^{\mathcal{I}}_t + \mathrm{MLP}_{\psi}(z^{\mathcal{I}}_t),
\label{eq:identity_adapter}
\end{equation}

where $\texttt{MLP}_{\psi}$ is initialized with zero. This ensures that at the start of transfer, $\tilde{z}^{\mathcal{I}}_0 = z^{\mathcal{I}}_0$, allowing the new agent to start with a valid protocol and gradually tune its signals to correct for cross-domain mismatches.

\paragraph{Summary.}
Overall, BRICKS-WM transforms the transfer learning challenge into a \emph{protocol-matching} problem: the target agent learns to control the environment by communicating with the frozen background transition module, enabling sample-efficient reuse of environmental dynamics in related settings. The reuse implementation details can refer to Appendix~\ref{app:warmup_and_reuse}.

\section{Experiments}

We begin by describing the experimental setup. Subsequently, we address the following research questions: (1) Can our framework achieve control performance comparable to monolithic state-of-the-art methods on standard locomotion benchmarks? (Section~\ref{exp:scratch_train_performance}); (2) Does the proposed framework effectively achieve dynamics decoupling and exhibit interpretable interface semantics?~(Section~\ref{exp:latent_modulation}); (3) Can the learned background dynamics be effectively reused across diverse transfer scenarios? (Section~\ref{exp:transfer}); (4) What is the impact of the interface regularization on the model's performance? (Section~\ref{exp:ablation}).

\textbf{Setup.} We evaluate BRICKS-WM on the DeepMind Control Suite (DMC)~\cite{dmc}. We select four diverse locomotion tasks: \emph{Walker Walk}, \emph{Walker Run}, \emph{Cheetah Run}, and \emph{Hopper Hop}. These tasks are chosen for their dynamic locomotion, which poses a significant challenge to the dynamics modeling capabilities of object-centric approaches. We compare against two state-of-the-art monolithic world models: \textbf{DreamerV3}~\cite{dreamerv3} and \textbf{TD-MPC2}~\cite{tdmpc2}. We focus our comparative analysis on these monolithic baselines, as prior object-centric approaches (e.g., SOLD~\cite{sold}) typically operate under a different setting requiring extensive pre-training or state supervision. In contrast, our goal is to verify that BRICKS-WM can be trained strictly \emph{from scratch} to achieve performance comparable to monolithic SOTA baselines (Section~\ref{exp:scratch_train_performance}), which subsequently serves for our transfer experiments (Section~\ref{exp:transfer}).

\subsection{Comparative Performance on Standard Locomotion Benchmarks}
\label{exp:scratch_train_performance}

Figure~\ref{fig:scratch_train_performance} presents the performance of various methods trained from scratch. Solid lines represent mean episode returns, while shaded areas represent the 95\% confidence intervals. Our approach achieves performance comparable to that of monolithic models, indicating that structural modularity can be achieved without sacrificing control efficacy. Notably, in the \emph{Hopper Hop} task, our model exhibits lower variance compared to baselines. We attribute this to the task's specific movement patterns: the hopping gait involves limited global displacement and less aggressive camera panning compared to high-speed running tasks like \emph{Cheetah Run}. This reduced visual variation simplifies the challenge of entity segmentation. Similar to the static manipulation tasks where object-centric approaches traditionally excel~\cite{sold, fiocwm}, the constrained background dynamics in \emph{Hopper Hop} facilitate robust structural disentanglement, thereby stabilizing policy optimization.

\begin{figure*}[tp!]
\centerline{\includegraphics[width=\linewidth]{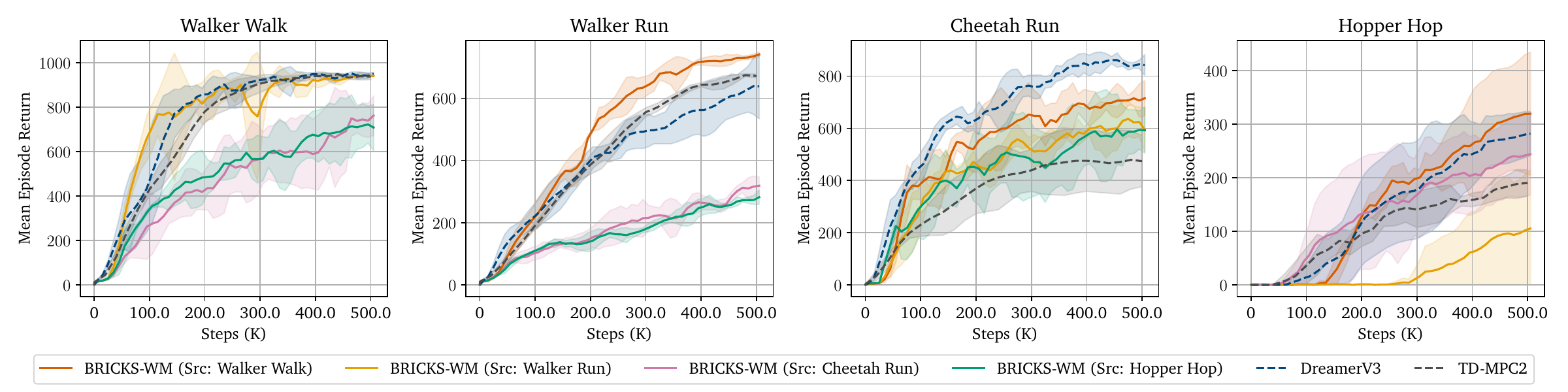}}
\caption{\textbf{Transfer performance on DMC.} We freeze a background module pre-trained on a Source task (e.g., \emph{Walker Walk}) and reuse it for a Target task (e.g., \emph{Walker Run}). The transfer agents (solid lines) often accelerate learning compared to training from scratch (dashed lines).}
\label{fig:transfer_performance}
\end{figure*}

\begin{figure*}[tp!]
\centerline{\includegraphics[width=\linewidth]{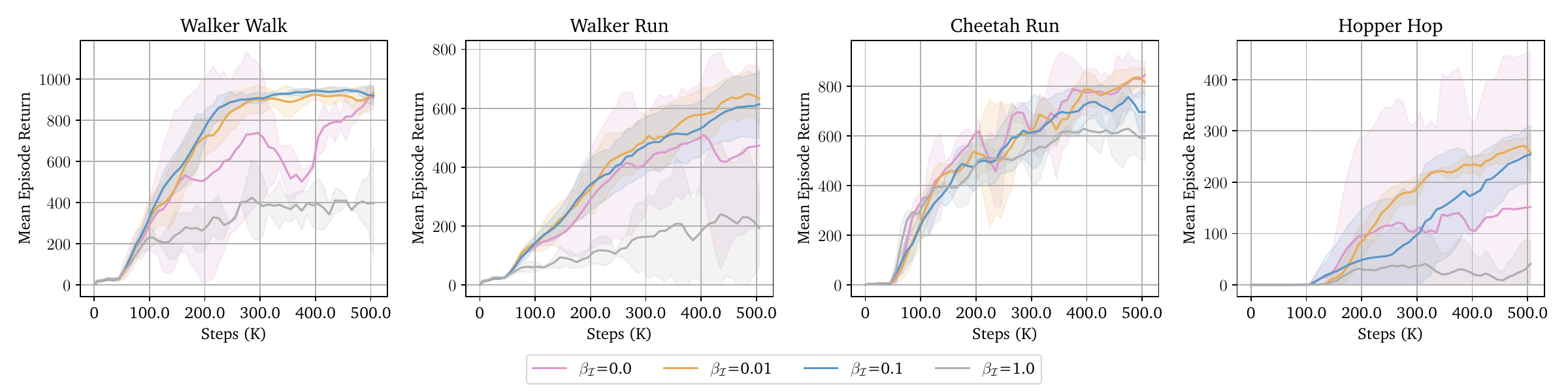}}
\caption{\textbf{Ablation study on the interface regularization coefficient.} A moderate penalty is crucial for establishing a standardized and effective interaction protocol.}
\label{fig:scratch_train_performance_ablation}
\end{figure*}

\subsection{Analyzing Interface Semantics and Dynamics Decoupling via Latent Modulation}
\label{exp:latent_modulation}

To interpret the semantic information captured by the latent variables and verify Assumption~\ref{ass:uni_interface}, we conduct a latent modulation experiment on \emph{Cheetah Run}. We sample an episode trajectory and employ the pre-trained model for observation reconstruction. During the generation process, we manually substitute the interface latent $z^{\mathcal{I}}$, originally produced by the agent interface policy, with a synthesized vector. Specifically, utilizing a 4-dimensional interface space, we modulate one dimension with a sinusoidal signal while fixing the remaining dimensions to zero. Figure~\ref{fig:visualization_cheetah_sine_wave} visualizes the results. We observe that the floor in the reconstructed observation exhibits vertical oscillation with the injected sine wave, while showing no horizontal displacement. This suggests that the modulated dimension selectively encodes the vertical offset of the background, thereby confirming that the interface semantics are functionally disentangled.

We further quantify this effect by comparing perturbed rollouts with their unperturbed baselines. For each interface dimension, we inject a sinusoidal perturbation during the first 50 imagined steps on 5 held-out episodes. A pixel is counted as changed if the mean absolute RGB difference exceeds 0.05. For the background region, we use the lower half of the frame, which captures the floor motion induced by locomotion.

\begin{table}[H]
\centering
\caption{\textbf{Quantitative interface perturbation analysis on \emph{Cheetah Run}.} BG and FG denote the background and foreground change rates, respectively.}

\label{tab:interface_perturbation}
{\small
\setlength{\tabcolsep}{8pt}
\renewcommand{\arraystretch}{1.16}
\begin{tabular}{lccc}
\toprule
\textbf{Dimension} & \textbf{BG} & \textbf{FG} & \textbf{BG/FG} \\
\midrule
Dim 0 & $0.156 \pm 0.005$ & $0.042 \pm 0.016$ & $3.727$ \\
Dim 1 & $0.171 \pm 0.010$ & $0.033 \pm 0.011$ & $5.206$ \\
Dim 2 & $0.166 \pm 0.010$ & $0.042 \pm 0.011$ & $3.951$ \\
Dim 3 & $0.165 \pm 0.010$ & $0.037 \pm 0.015$ & $4.413$ \\
\bottomrule
\end{tabular}
}

\end{table}

As shown in Table~\ref{tab:interface_perturbation}, perturbing the interface changes the background 3.7--5.2 times more frequently than the foreground. This quantitative result corroborates the qualitative visualization, indicating that the learned interface primarily modulates background dynamics rather than leaking detailed agent appearance.

\subsection{Transfer via Module Reuse}
\label{exp:transfer}

\begin{table}[H]
\centering
\caption{\textbf{Final return at 500K target-environment steps.} Rows denote source tasks and columns denote target tasks for transfer.}

\label{tab:transfer_final_return}
{\small
\setlength{\tabcolsep}{6pt}
\renewcommand{\arraystretch}{1.16}
\begin{tabular}{lcccc}
\toprule
\textbf{Source $\backslash$ Target} & \textbf{Walker Walk} & \textbf{Walker Run} & \textbf{Cheetah Run} & \textbf{Hopper Hop} \\
\midrule
Walker Walk & -- & $746 \pm 4$ & $722 \pm 78$ & $321 \pm 100$ \\
Walker Run & $940 \pm 9$ & -- & $566 \pm 112$ & $97 \pm 88$ \\
Cheetah Run & $780 \pm 37$ & $318 \pm 23$ & -- & $249 \pm 69$ \\
Hopper Hop & $653 \pm 124$ & $309 \pm 22$ & $598 \pm 99$ & -- \\
\bottomrule
\end{tabular}
}

\end{table}

\begin{table}[H]
\centering
\caption{\textbf{Final AUC at 500K target-environment steps.} Rows denote source tasks and columns denote target tasks for transfer. AUC is timestep-normalized.}

\label{tab:transfer_final_auc}
{\small
\setlength{\tabcolsep}{6pt}
\renewcommand{\arraystretch}{1.16}
\begin{tabular}{lcccc}
\toprule
\textbf{Source $\backslash$ Target} & \textbf{Walker Walk} & \textbf{Walker Run} & \textbf{Cheetah Run} & \textbf{Hopper Hop} \\
\midrule
Walker Walk & -- & $508 \pm 18$ & $538 \pm 48$ & $158 \pm 31$ \\
Walker Run & $776 \pm 44$ & -- & $454 \pm 34$ & $28 \pm 29$ \\
Cheetah Run & $477 \pm 67$ & $186 \pm 17$ & -- & $143 \pm 62$ \\
Hopper Hop & $506 \pm 59$ & $173 \pm 9$ & $435 \pm 78$ & -- \\
\bottomrule
\end{tabular}
}

\end{table}

\begin{table}[H]
\centering
\caption{\textbf{Scratch baselines at 500K environment steps.} Each entry reports final return and final AUC, respectively.}

\label{tab:scratch_baselines_500k}
{\footnotesize
\setlength{\tabcolsep}{3pt}
\renewcommand{\arraystretch}{1.16}
\begin{tabular}{lcccc}
\toprule
\textbf{Method} & \textbf{Walker Walk} & \textbf{Walker Run} & \textbf{Cheetah Run} & \textbf{Hopper Hop} \\
\midrule
DreamerV3 & $958 \pm 12 / 774 \pm 50$ & $636 \pm 113 / 421 \pm 30$ & $821 \pm 56 / 662 \pm 16$ & $281 \pm 42 / 147 \pm 56$ \\
TD-MPC2 & $952 \pm 10 / 741 \pm 21$ & $662 \pm 20 / 449 \pm 9$ & $455 \pm 58 / 372 \pm 68$ & $194 \pm 20 / 114 \pm 17$ \\
\bottomrule
\end{tabular}
}

\end{table}

A distinctive capability of BRICKS-WM is the modular reusability of the background dynamics. We evaluate the reusability by transferring the frozen background dynamics module from a \emph{Source} to a \emph{Target} task, training agent-specific components from scratch. Figure~\ref{fig:transfer_performance} presents results for both \emph{in-domain transfer} (consistent physics) and scenarios involving \emph{domain shifts}. For in-domain tasks (i.e., \emph{Walker Walk} $\leftrightarrow$ \emph{Walker Run}), bidirectional transfer matches or exceeds strong monolithic baselines. The observed rapid initial convergence demonstrates that the background dynamics module reuse simplifies the learning problem to protocol alignment. Beyond in-domain reuse, the framework exhibits potential under moderate domain shifts. For instance, transferring \emph{Walker Walk} background dynamics to the morphologically distinct \emph{Hopper Hop} task yields returns surpassing the baselines, suggesting possible adaptability to distributional shifts. Conversely, scenarios where transfer performance lags behind the baselines under significant domain gaps serve to empirically delineate the boundaries of effective reuse. Collectively, these results position BRICKS-WM as a competitive approach. Tables~\ref{tab:transfer_final_return}, \ref{tab:transfer_final_auc}, and~\ref{tab:scratch_baselines_500k} summarize the final return, timestep-normalized AUC at 500K target-environment steps, and the scratch baselines, respectively. Table~\ref{tab:transfer_final_return} shows that in-domain reuse is competitive with scratch training, with \emph{Walker Walk} $\rightarrow$ \emph{Walker Run} reaching a final return of $746 \pm 4$ and \emph{Walker Run} $\rightarrow$ \emph{Walker Walk} reaching $940 \pm 9$. Table~\ref{tab:transfer_final_auc} further indicates consistent sample-efficiency gains in the same directions, while comparison with Table~\ref{tab:scratch_baselines_500k} shows that the benefit becomes more task-dependent under larger domain shifts.

\subsection{Ablation Study}
\label{exp:ablation}

We analyze the necessity of the interface regularization mechanism, $\mathcal{R}_{\mathcal{I}}$, by varying its coefficient $\beta_{\mathcal{I}}$. This term acts as an information bottleneck intended to enforce a standardized interaction protocol. Figure~\ref{fig:scratch_train_performance_ablation} compares the performance. When $\beta_{\mathcal{I}}=0$, the training exhibits substantial variance and instability. This suggests that without the information bottleneck constraint, the interface fails to filter out agent-specific details. This leakage undermines the intended modular decoupling. Conversely, an overly strong regularization (e.g., $\beta_{\mathcal{I}}=1.0$) excessively restricts the interface capacity. This prevents the model from encoding high-frequency physical quantities required for dynamic locomotion, leading to performance degradation in high-dynamic tasks like \emph{Cheetah Run} and \emph{Walker Run}. Consequently, we adopt $\beta_{\mathcal{I}}=0.01$ as the optimal setting, striking a balance between enforcing an agent-agnostic protocol and maintaining sufficient information for control.

\section{Conclusion and Future Work}
In this paper, we introduce BRICKS-WM, a framework designed to decouple global dynamics into a composition of distinct, reusable modules interacting via learned latent interfaces. As a minimal instantiation, we implement a functional decomposition of the world into an actuated Agent module and an external Background module, ensuring that the background dynamics remains agnostic to the agent. Empirically, to the best of our knowledge, BRICKS-WM is the first object-centric world model trained from scratch to achieve control performance on locomotion tasks comparable to strong monolithic baselines. Furthermore, BRICKS-WM achieves competitive performance by reusing the frozen background dynamics, turning the transfer problem into a protocol matching problem. While currently validated through an Agent-Background decomposition, the BRICKS-WM framework allows for broader modularity. The current study is limited to two-slot DMC settings, and richer bidirectional interactions, cluttered multi-object scenes, and non-reconstruction objectives remain important future directions. Future work will leverage advanced visual representations to scale to complex tasks, enabling fine-grained dynamics decomposition and the scalable assembly of complex environments.






\bibliography{bricks_wm}
\bibliographystyle{icml2026}

\newpage
\appendix
\onecolumn

\section{Derivation of the Evidence Lower Bound}
\label{app:elbo}

In this section, we provide the detailed derivation of the Evidence Lower Bound (ELBO) objective used in Sec.~\ref{sec:training_obj}.
An episode is denoted by $\{(o_t,a_t)\}_{t=1}^{T}$ and we condition on the action sequence. We follow the indexing convention where latent states $s_t$ generate observations $o_t$, and the transition from $t-1$ to $t$ is driven by action $a_{t-1}$.

\subsection{Generative Model and Factorization}

We consider an episode of length $T$. The model introduces three sets of latent variables: Agent states $s^{ag}_{1:T}$, Background states $s^{bg}_{1:T}$, and Interface latents $z^{\mathcal{I}}_{0:T-1}$ (where $z^{\mathcal{I}}_{t-1}$ mediates the transition from $t-1$ to $t$).

We define fixed \emph{dummy} initial conditions $s^{ag}_0, s^{bg}_0$. The probabilistic dependence at $t=1$ is expressed as a transition from these dummy states (e.g., $p(s^{ag}_1) \triangleq p(s^{ag}_1 | s^{ag}_0, a_0)$).

Based on the modular factorization described in Eq.~\ref{eq:dynamics_factorization}, the joint distribution is

\begin{equation}
\begin{aligned}
p_{\theta}(o_{1:T}, s^{ag}_{1:T}, s^{bg}_{1:T}, z^{\mathcal{I}}_{0:T-1} \mid a_{1:T})
    &= \prod_{t=1}^{T} \bigg[
    \underbrace{p_{\theta_o}(o_t \mid s^{ag}_t, s^{bg}_t)}_{\text{Observation Decoder}} \\
    &\quad \cdot
    \underbrace{p_{\theta_{ag}}(s^{ag}_{t} \mid s^{ag}_{t-1}, a_{t-1})}_{\text{Agent Dynamics}} \cdot
    \underbrace{p_{\theta_{\mathcal{I}}}(z^{\mathcal{I}}_{t-1} \mid s^{ag}_{t}, a_{t-1})}_{\text{Interface Policy}} \cdot
    \underbrace{p_{\theta_{bg}}(s^{bg}_{t} \mid s^{bg}_{t-1}, z^{\mathcal{I}}_{t-1})}_{\text{Background Dynamics}}
    \bigg].
\end{aligned}
\label{eq:app_joint}
\end{equation}

Note that for $t=1$, the interface term refers to the initial dummy latent $z^{\mathcal{I}}_0$.

\subsection{Variational Inference and ELBO Decomposition}
The inference model $q_{\phi}$ approximates the posterior using the current observation $o_t$. Consistent with the generative structure, the interface latent is inferred via the agent's transition:

\begin{equation}
\begin{aligned}
q_{\phi}(s^{ag}_{1:T}, s^{bg}_{1:T}, z^{\mathcal{I}}_{0:T-1} \mid o_{1:T}, a_{1:T})
    &= \prod_{t=1}^{T} \bigg[
    q_{\phi_{ag}}(s^{ag}_{t} \mid s^{ag}_{t-1}, a_{t-1}, o_t) \\
    &\quad \cdot q_{\phi_{\mathcal{I}}}(z^{\mathcal{I}}_{t-1} \mid s^{ag}_{t}, a_{t-1})
    \cdot q_{\phi_{bg}}(s^{bg}_{t} \mid s^{bg}_{t-1}, z^{\mathcal{I}}_{t-1}, o_t)
    \bigg].
\end{aligned}
\label{eq:app_posterior}
\end{equation}

Starting from the marginal likelihood,
\begin{align}
\log p_{\theta}(o_{1:T}\mid a_{1:T})
&=
\log \int
p_{\theta}(o_{1:T}, s^{ag}_{1:T}, s^{bg}_{1:T}, z^{\mathcal{I}}_{0:T-1}\mid a_{1:T})
\, d s^{ag}_{1:T}\, d s^{bg}_{1:T}\, d z^{\mathcal{I}}_{0:T-1}
\nonumber\\
&=
\log \mathbb{E}_{q_{\phi}}
\left[
\frac{
p_{\theta}(o_{1:T}, s^{ag}_{1:T}, s^{bg}_{1:T}, z^{\mathcal{I}}_{0:T-1}\mid a_{1:T})
}{
q_{\phi}(s^{ag}_{1:T}, s^{bg}_{1:T}, z^{\mathcal{I}}_{0:T-1}\mid o_{1:T}, a_{1:T})
}
\right].
\label{eq:app_is}
\end{align}

Applying Jensen's inequality yields the ELBO:
\begin{align}
\log p_{\theta}(o_{1:T}\mid a_{1:T})
\ge
\mathbb{E}_{q_{\phi}}
\left[
\log p_{\theta}(o_{1:T}, s^{ag}_{1:T}, s^{bg}_{1:T}, z^{\mathcal{I}}_{0:T-1}\mid a_{1:T})
-
\log q_{\phi}(s^{ag}_{1:T}, s^{bg}_{1:T}, z^{\mathcal{I}}_{0:T-1}\mid o_{1:T}, a_{1:T})
\right].
\label{eq:app_jensen}
\end{align}

Substituting the factorizations in Eq.~\ref{eq:app_joint} and Eq.~\ref{eq:app_posterior} into Eq.~\ref{eq:app_jensen}, we decompose the ELBO into reconstruction terms and KL divergences
\begin{equation}
\label{eq:app_expand}
\begin{split}
\mathcal{L}_{\textsc{elbo}}
&=
\sum_{t=1}^{T}\mathbb{E}_{q_{\phi}}\!\left[\log p_{\theta_o}(o_t\mid s^{ag}_{t}, s^{bg}_{t})\right] \\
&\quad - \sum_{t=1}^{T}\mathbb{E}_{q_{\phi}}\!\left[
\log \frac{q_{\phi_{ag}}(s^{ag}_{t}\mid s^{ag}_{t-1}, a_{t-1}, o_t)}
{p_{\theta_{ag}}(s^{ag}_{t}\mid s^{ag}_{t-1}, a_{t-1})}
\right] \\
&\quad - \sum_{t=1}^{T}\mathbb{E}_{q_{\phi}}\!\left[
\log \frac{q_{\phi_{bg}}(s^{bg}_{t}\mid s^{bg}_{t-1}, z^{\mathcal{I}}_{t-1}, o_t)}
{p_{\theta_{bg}}(s^{bg}_{t}\mid s^{bg}_{t-1}, z^{\mathcal{I}}_{t-1})}
\right] \\
&\quad - \sum_{t=1}^{T}\mathbb{E}_{q_{\phi}}\!\left[
\log \frac{q_{\phi_{\mathcal{I}}}(z^{\mathcal{I}}_{t-1}\mid s^{ag}_{t}, a_{t-1})}
{p_{\theta_{\mathcal{I}}}(z^{\mathcal{I}}_{t-1}\mid s^{ag}_{t}, a_{t-1})}
\right].
\end{split}
\end{equation}

\paragraph{Cancellation of the Interface KL Term.}
Consider the specific KL divergence term associated with the interface latent $z^{\mathcal{I}}$. In our framework, the interface is an intermediate variable bridging the Agent and Background modules. During training, we define the variational posterior $q_{\phi_{\mathcal{I}}}$ to share the same parameterization as the generative policy $p_{\theta_{\mathcal{I}}}$ (both conditioned on the agent's kinematic state):
\begin{equation}
    q_{\phi_{\mathcal{I}}}(z^{\mathcal{I}}_{t-1} \mid s^{ag}_{t}, a_{t-1}) \equiv p_{\theta_{\mathcal{I}}}(z^{\mathcal{I}}_{t-1} \mid s^{ag}_{t}, a_{t-1}).
\end{equation}
Consequently, the log-ratio of these distributions is zero, and the KL term vanishes from the standard ELBO:
\begin{equation}
\mathbb{E}_{q_{\phi}}\!\left[
\log \frac{q_{\phi_{\mathcal{I}}}(z^{\mathcal{I}}_{t-1}\mid s^{ag}_{t}, a_{t-1})}
{p_{\theta_{\mathcal{I}}}(z^{\mathcal{I}}_{t-1}\mid s^{ag}_{t}, a_{t-1})}
\right]
=
0.
\end{equation}
This derivation confirms why the interface variable $z^{\mathcal{I}}$ does not appear in the dynamics regularization term $\mathcal{R}_{\textsc{kl}}$ (Eq.~\ref{eq:bricks_kl_decompose}) in the main text.

Therefore, the ELBO contains \textbf{no} KL term for $z^{\mathcal{I}}$, consistent with the main-text objective where $z^{\mathcal{I}}$ is treated as an intermediate interaction/protocol variable.

The reconstruction term can be written as:
\begin{equation}
\label{eq:app_rec}
\mathcal{L}_{\textsc{rec}}
=
\sum_{t=1}^{T}\mathbb{E}_{q_{\phi}}\!\left[\log p_{\theta_o}(o_t\mid s^{ag}_{t}, s^{bg}_{t})\right].
\end{equation}

The remaining ratio terms can be written as KL divergences:
\begin{equation}
\label{eq:app_kl_ag}
\mathcal{L}_{\textsc{kl}}^{ag}
=
\sum_{t=1}^{T}
\mathbb{E}_{q_{\phi}}\!\left[
\KL\!\Big(
q_{\phi_{ag}}(s^{ag}_{t}\mid s^{ag}_{t-1}, a_{t-1}, o_t)
\,\Big\|\,
p_{\theta_{ag}}(s^{ag}_{t}\mid s^{ag}_{t-1}, a_{t-1})
\Big)
\right],
\end{equation}
\begin{equation}
\label{eq:app_kl_bg}
\mathcal{L}_{\textsc{kl}}^{bg}
=
\sum_{t=1}^{T}
\mathbb{E}_{q_{\phi}}\!\left[
\KL\!\Big(
q_{\phi_{bg}}(s^{bg}_{t}\mid s^{bg}_{t-1}, z^{\mathcal{I}}_{t-1}, o_t)
\,\Big\|\,
p_{\theta_{bg}}(s^{bg}_{t}\mid s^{bg}_{t-1}, z^{\mathcal{I}}_{t-1})
\Big)
\right].
\end{equation}

Combining Eq.~\ref{eq:app_rec}--\ref{eq:app_kl_bg}, we obtain the final ELBO:
\begin{equation}
\label{eq:app_elbo_final}
\mathcal{L}_{\textsc{elbo}}
=
\mathcal{L}_{\textsc{rec}}
-
\left(\mathcal{L}_{\textsc{kl}}^{ag}+\mathcal{L}_{\textsc{kl}}^{bg}\right),
\end{equation}
which corresponds to Eq.~\ref{eq:bricks_elbo_main} and Eq.~\ref{eq:bricks_kl_decompose} in Sec.~\ref{sec:training_obj}.

\section{Pseudo Code}

The complete training procedure for the source task is presented in Algorithm~\ref{alg:brickswm}, while the dynamics reuse mechanism for the target task is detailed in Algorithm~\ref{alg:brickswm_reuse}. For brevity, we omit predictors for other task information, retaining only the reward predictor.

\begin{algorithm}[hp]
    \caption{BRICKS-WM Training (Source Task)}
    \label{alg:brickswm}
\begin{algorithmic}
    \STATE {\bfseries Initialize:} Dataset $\mathcal{D}$ collected by random policy, World Model parameters $\theta$ (comprising Agent $\theta_{ag}$, Background $\theta_{bg}$, Interface $\theta_{\mathcal{I}}$), Encoder parameters $\phi$ (Slot Attention), Actor parameters $\psi$, and Critic parameters $\xi$
    \FOR{training step $t_1=1...T_1$}
        \FOR{update step $t_2=1...T_2$}
        \STATE // \texttt{Dynamics \& Interface Learning}
        \STATE Sample minibatch sequence $\{(o_t, a_t, r_t) \}_{t=k}^{k+L} \sim \mathcal{D}$
        \STATE \textbf{Perception:} Infer factorized posteriors $(s_t^{ag}, s_t^{bg}) \sim q_\phi(s_t|o_t)$ via Modular Encoder
        \STATE \textbf{Generation:}
        \STATE \quad Compute Agent prior $s_t^{ag} \sim p_{\theta_{ag}}(s_t^{ag}|s_{t-1}^{ag}, a_{t-1})$
        \STATE \quad Generate Interface Latent $z_{t-1}^{\mathcal{I}} \sim p_{\theta_{\mathcal{I}}}(z_{t-1}^{\mathcal{I}}|s_t^{ag}, a_{t-1})$
        \STATE \quad Compute Background prior $s_t^{bg} \sim p_{\theta_{bg}}(s_t^{bg}|s_{t-1}^{bg}, z_{t-1}^{\mathcal{I}})$
        \STATE \textbf{Reconstruction:}
        \STATE \quad Decode Agent components $(\hat{o}_t^{ag}, m_t) = D_{\theta_{ag}}(s_t^{ag})$ (RGB and Alpha Mask)
        \STATE \quad Decode Background $\hat{o}_t^{bg} = D_{\theta_{bg}}(s_t^{bg})$
        \STATE \quad Composite image $\hat{o}_t = m_t \odot \hat{o}_t^{ag} + (1-m_t) \odot \hat{o}_t^{bg}$
        \STATE \textbf{Optimization:}
        \STATE \quad Calculate $\mathcal{L}_{ELBO}$ (Reconstruction and Dynamics KL)
        \STATE \quad Calculate Regularizations: Interface $\mathcal{R}_{\mathcal{I}}$  and Mask Sparsity $\mathcal{R}_{mask}$
        \STATE \quad Update parameters $\theta, \phi$ on Loss $\mathcal{L} = -\mathcal{L}_{ELBO} + \beta_{mask}\mathcal{R}_{mask} + \beta_{\mathcal{I}}\mathcal{R}_{\mathcal{I}}$

        \STATE // \texttt{Behavior Learning}
        \STATE \textbf{Imagination:} Rollout trajectories $\{(s_\tau, a_\tau, r_\tau) \}_{\tau=t}^{t+H}$ from current posteriors:
        \STATE \quad Sample action $a_\tau \sim \pi_\psi(a_\tau | s_\tau^{ag}, s_\tau^{bg})$
        \STATE \quad Step Modular Dynamics: $s_\tau^{ag} \xrightarrow{p_{\theta_{ag}}} s_{\tau+1}^{ag} \xrightarrow{p_{\theta_{\mathcal{I}}}} z_{\tau}^{\mathcal{I}} \xrightarrow{p_{\theta_{bg}}} s_{\tau+1}^{bg}$
        \STATE \quad Predict reward $\hat{r}_\tau$
        \STATE Update Actor $\psi$ and Critic $\xi$ using imagined returns
        \ENDFOR

        \FOR{rollout step $t_3=1...T_3$}
            \STATE Observe $o_t$
            \STATE Infer factorized state $(s_t^{ag}, s_t^{bg}) \sim q_\phi(s_t|o_t)$
            \STATE Sample action from exploration policy $a_t \sim \pi_{\psi}(a_t|s_t^{ag}, s_t^{bg})$
            \STATE Execute action: $r_t, o_{t+1} \leftarrow$ \texttt{env.step($a_t$)}
            \STATE Add experience to dataset $\mathcal{D} \leftarrow \mathcal{D} \cup \{(o_t, a_t, r_t)\} $
        \ENDFOR
    \ENDFOR
\end{algorithmic}
\end{algorithm}

\begin{algorithm}[tb]
    \caption{BRICKS-WM Dynamics Reuse (Target Task)}
    \label{alg:brickswm_reuse}
\begin{algorithmic}
    \STATE {\bfseries Input:} Source parameters $\theta^{src}$, Target Dataset $\mathcal{D}_{tgt}$
    \STATE {\bfseries Transfer Setup:}
    \STATE \quad 1. \textbf{Inherit:} $\theta_{bg}, \theta_{enc}, \theta_{dec} \leftarrow \theta^{src}$ \quad \textcolor{black}{// Load BG Dynamics, Encoder, Decoder}
    \STATE \quad 2. \textbf{Partial Freeze:}
    \STATE \quad \quad \textbf{Freeze} $\theta_{enc}$ (Spatial \& Slot Attn weights), $\theta_{dec}$.
    \STATE \quad \quad \textbf{Freeze} $\theta_{bg}^{core}$ (RNN Cell \& Input Layers).
    \STATE \quad \quad \textbf{Unfreeze} $\theta_{bg}^{heads}$ (Output Layers) for domain adaptation.
    \STATE \quad 3. \textbf{Re-initialize:} $\theta_{ag}$ (Agent Dynamics), $\theta_{\mathcal{I}}$ (Interface), $\phi_{ag\_init}$ (Agent Slot Query).
    \STATE \quad 4. \textbf{Add Adapter:} Initialize Interface Adapter $A_\psi$ (as identity).

    \FOR{training step $t=1...T$}
        \STATE // \texttt{Inference \& Reconstruction}
        \STATE Sample batch $o_{t}$ from $\mathcal{D}_{tgt}$
        \STATE $(s_t^{ag}, s_t^{bg}) \sim q_{\phi}(s_t|o_t)$ \quad \textcolor{black}{// Encoder uses re-inited Agent Query}

        \STATE // \texttt{Modified Dynamics Step with Adapter}
        \STATE \textbf{Generation (Re-assembled):}
        \STATE \quad $s_t^{ag} \sim p_{\theta_{ag}}(s_t^{ag}|s_{t-1}^{ag}, a_{t-1})$ \quad \textcolor{black}{// Learned Agent Dynamics}
        \STATE \quad $z_{t-1}^{\mathcal{I}} \sim p_{\theta_{\mathcal{I}}}(z_{t-1}^{\mathcal{I}}|s_t^{ag}, a_{t-1})$ \quad \textcolor{black}{// New Interface Policy}
        \STATE \quad $\tilde{z}_{t-1}^{\mathcal{I}} = A_\psi(z_{t-1}^{\mathcal{I}}) = z_{t-1}^{\mathcal{I}} + \text{MLP}(z_{t-1}^{\mathcal{I}})$ \quad \textcolor{black}{// Interface Adapter}
        \STATE \quad $s_t^{bg} \sim p_{\theta_{bg}}(s_t^{bg}|s_{t-1}^{bg}, \tilde{z}_{t-1}^{\mathcal{I}})$ \quad \textcolor{black}{// Frozen Core, Adapted Heads}

        \STATE \textbf{Optimization:}
        \STATE \quad Update $\theta_{ag}, \theta_{\mathcal{I}}, \phi_{ag\_init}, A_\psi, \theta_{bg}^{heads}$ to minimize $\mathcal{L}$
    \ENDFOR
\end{algorithmic}
\end{algorithm}

\section{Implementation Details}
\label{app:implement_detail}

\subsection{Modular Encoder Details}
\label{app:modular_encoder_datail}

To functionally decouple the visual scene into entity-centric representations, we employ a Modular Encoder composed of a Convolutional Neural Network (CNN) backbone followed by a Slot Attention module.

\paragraph{Convolutional Backbone.}
Given an input observation $\mathbf{o}_t \in \mathbb{R}^{H \times W \times 3}$, we first extract a feature map using a CNN encoder. Unlike standard encoders that flatten the spatial dimensions immediately, we preserve the spatial structure to allow for object-centric binding. The encoder produces a feature map $\mathbf{E}_{\text{spatial}} \in \mathbb{R}^{H' \times W' \times D_{\text{enc}}}$. We flatten the spatial dimensions to obtain a set of $N = H' \times W'$ input feature vectors $\mathbf{V}_{\text{in}} \in \mathbb{R}^{N \times D_{\text{enc}}}$, which serve as the input to the attention mechanism.

\paragraph{Slot Attention.}
We adapt the Slot Attention mechanism~\citep{slot_attention} to map the $N$ input vectors to $N_s=2$ object slots: the Agent slot $s^{\text{ag}}$ and the Background slot $s^{\text{bg}}$. Standard Slot Attention initializes slots by sampling from a shared Gaussian distribution, which leads to permutation equivariant slots (order does not matter). However, for BRICKS-WM, we require semantic binding (i.e., the first slot must always represent the Agent and the second the Background) to facilitate modular reuse. Therefore, instead of random sampling, we define the initial slots $\mathbf{S}_0 \in \mathbb{R}^{N_s \times D_{\text{slot}}}$ as distinct learnable parameters:
\begin{align}
    \mathbf{S}_0 = \begin{bmatrix} \boldsymbol{\mu}_{\text{ag}} \\ \boldsymbol{\mu}_{\text{bg}} \end{bmatrix},
\end{align}
where $\boldsymbol{\mu}_{\text{ag}}, \boldsymbol{\mu}_{\text{bg}} \in \mathbb{R}^{D_{\text{slot}}}$ are independent learnable vectors initialized with Gaussian noise. The slots are refined iteratively over $T_{\text{iter}}$ iterations (set to 3 in our experiments) via a competitive attention mechanism. At each iteration $l=1, \dots, T_{\text{iter}}$, the update proceeds as follows:

\begin{enumerate}
    \item \textbf{Projection:} We project the inputs $\mathbf{V}_{\text{in}}$ into keys $\mathbf{K}$ and values $\mathbf{V}$, and the layer-normalized slots $\mathbf{S}_{l-1}$ into queries $\mathbf{Q}$:
        \begin{align}
        \mathbf{K} = \text{LayerNorm}(\mathbf{V}_{in})\mathbf{W}_k, \quad
        \mathbf{V} = \text{LayerNorm}(\mathbf{V}_{in})\mathbf{W}_v, \quad
        \mathbf{Q} = \text{LayerNorm}(\mathbf{S}_{l-1})\mathbf{W}_q.
    \end{align}
    where $\mathbf{W}_q, \mathbf{W}_k, \mathbf{W}_v$ are learnable linear projections.

    \item \textbf{Attention:} We compute the dot-product attention scores. Crucially, we apply the softmax function over the \emph{slots} dimension $N_s$. This forces the slots to compete for explaining each part of the input image:
    \begin{align}
        \mathbf{M} &= \frac{1}{\sqrt{D_{\text{slot}}}} \mathbf{Q} \mathbf{K}^\top \in \mathbb{R}^{N_s \times N}, \\
        \alpha_{i,j} &= \frac{\exp(M_{i,j})}{\sum_{k=1}^{N_s} \exp(M_{k,j})},
    \end{align}
    where $\alpha_{i,j}$ represents the attention weight of slot $i$ for input feature vector $j$.

    \item \textbf{Aggregation:} To ensure the update magnitude is independent of the object size (number of pixels), we normalize the attention weights over the spatial dimension $N$ before aggregating the values:
    \begin{align}
        w_{i,j} &= \frac{\alpha_{i,j}}{\sum_{p=1}^{N} \alpha_{i,p} + \epsilon}, \\
        \mathbf{U}_i &= \sum_{j=1}^{N} w_{i,j} \mathbf{V}_j.
    \end{align}
    Here, $\mathbf{U}_i$ is the aggregated update for the $i$-th slot.

    \item \textbf{Update:} The aggregated updates $\mathbf{U}$ are used to update the slots using a Gated Recurrent Unit (GRU), followed by an MLP with a residual connection:
    \begin{align}
        \hat{\mathbf{S}}_l &= \text{GRU}(\mathbf{U}, \mathbf{S}_{l-1}), \\
        \mathbf{S}_l &= \hat{\mathbf{S}}_l + \text{MLP}(\text{LayerNorm}(\hat{\mathbf{S}}_l)).
    \end{align}
    Note that the GRU parameters are shared across all slots.
\end{enumerate}

After $T_{\text{iter}}$ iterations, the final slots are projected via a linear layer to match the dimension required by the dynamics modules. Finally these dynamics modules yield the factorized posterior state inputs $s^{ag}_t$ and $s^{bg}_t$.

\subsection{Warmup and Reuse Details}
\label{app:warmup_and_reuse}

\textbf{Warmup Phase.}
To facilitate the stable emergence of object-centric representations before imposing dynamics constraints, we employ a warmup strategy during the initial phase of training. Specifically, for the first $N_{\texttt{warmup}}$ updates  (set to 10,000 for source tasks and 5,000 for target tasks), we suppress the dynamics consistency loss and the interface regularization. We set the coefficients $\beta_{dyn} = \beta_{rep} = \beta_{\mathcal{I}} = 0$, reducing the objective to pure reconstruction maximization: $\mathcal{L} \approx \mathcal{L}_{REC} + \beta_{mask}\mathcal{R}_{mask}$. This allows the Slot Attention mechanism to function as an autoencoder initially, ensuring a good entity segmentation initialization before the model attempts to learn the temporal transition dynamics.

\textbf{Dynamics Reuse and Freezing Strategy.}
When transferring the background module to a new agent, we strictly partition the model parameters into inherited, frozen, and re-initialized sets to enforce the reuse protocol:

\begin{itemize}
    \item \textbf{Parameter Inheritance:} We initialize the target model by loading the pre-trained parameters $\theta_{src}$ for the domain-invariant components: the Spatial Encoder (CNN), the core of Slot Attention (Keys, Values, GRU, and Background Query), the Background Dynamics model ($\theta_{bg}$), and the Decoder.

    \item \textbf{Re-initialization:} Agent-specific components are initialized from scratch. These include the Agent Query in Slot Attention ($\phi_{ag\_init}$), the Agent Dynamics model ($\theta_{ag}$), the Interface Policy ($\theta_{\mathcal{I}}$), and the Task Heads (Reward and Continuation).

\item \textbf{Perceptual Freezing and Slot Adaptation:}
During training on the target task, we freeze the weights of the Spatial Encoder to preserve the visual feature space. For the Slot Attention module, we freeze its shared extraction mechanisms—specifically the linear projections for queries, keys, and values, the recurrent update function (GRU), and the feedforward networks (MLP). This ensures the object-centric binding logic remains consistent. Crucially, while the learnable background slot query ($\phi_{bg}$) is frozen to anchor the scene representation, the agent slot query ($\phi_{ag}$) is re-initialized and trainable, enabling the model to learn to bind to the new agent using the pre-established attention mechanism.

    \item \textbf{Core Dynamics Freezing:}
    For the Background Dynamics, we enforce structural consistency by freezing the recurrent GRU cells and input projection layers (core Background Dynamics). However, we allow the RSSM output layers to be fine-tuned to adapt to domain shifts. Additionally, a zero-initialized residual adapter (MLP) is inserted before the background module to gradually align the interface distribution $\tilde{z}^{\mathcal{I}}$ without disrupting the initial transfer.

\end{itemize}

\subsection{Hyperparameters and Time Cost}
\label{app:hyper_time_cost}

Table~\ref{tab:hyperparameters} presents the primary hyperparameters of BRICKS-WM. Our implementation is built upon the DreamerV3 codebase~\footnote{https://github.com/NM512/dreamerv3-torch}. The Agent and Background modules are instantiated as two separate Recurrent State-Space Models (RSSMs) but share the same structural hyperparameters (e.g., recurrent units and latent dimensions) as listed in the table. The behavior policy operates on the concatenated features extracted from both the Agent and Background modules ($s_t = [s_t^{ag}, s_t^{bg}]$) to ensure the agent can react to environmental changes. The experiments are conducted on NVIDIA RTX 4090 GPUs. With a batch size of 16 and sequence length of 64, training on DMC tasks requires approximately 24 hours for 500K environment steps.

\begin{table}[ht!]
\centering
\caption{Hyperparameters for BRICKS-WM}
\begin{tabular}{lc}
\toprule
 \textbf{Hyperparameter} & \textbf{Value} \\
\midrule
\multicolumn{2}{l}{\textit{General}} \\
Action Repeat & 2 \\
Batch Size & 16 \\
Batch Length & 64 \\
Optimizer & Adam \\
Learning Rate (Model) & $1 \times 10^{-4}$ \\
Learning Rate (Actor/Critic) & $3 \times 10^{-5}$ \\
\midrule
\multicolumn{2}{l}{\textit{Loss Scales}} \\
Dynamics Scale $\beta_{\mathrm{dyn}}$ & 0.5 \\
Representation Scale $\beta_{\mathrm{rep}}$ & 0.1 \\
Interface KL Scale $\beta_{\mathcal{I}}$ & 0.01 \\
Mask Loss Scale $\beta_{\mathrm{mask}}$ & 1.0 \\
Free Bits (KL) & 1.0 \\
\midrule
\multicolumn{2}{l}{\textit{Modular RSSM (Agent $s_t^{ag}$ \& Background $s_t^{bg}$)}} \\
Discrete latent dimensions & 32 \\
Discrete latent classes & 32 \\
GRU recurrent units (Deterministic) & 512 \\
Dense hidden units & 512 \\
Activation & SiLU \\
Normalization & LayerNorm \\
\midrule
\multicolumn{2}{l}{\textit{Interface \& Attention}} \\
Interface latent dimension ($z^\mathcal{I}$) & 4 \\
Slot Number & 2 (Agent + Background) \\
Slot Dimension & 64 \\
Slot Attention Iterations $T_{\texttt{iter}}$ & 3 \\
Warmup Update Times $N_{\texttt{warmup}}$ & 10,000 for source, 5,000 for reuse \\
\bottomrule
\end{tabular}
\label{tab:hyperparameters}
\end{table}

\section{Additional Experiment Results}
\label{app:more_exps}

\subsection{Perception-Only Warm Start Control}
\label{app:perception_only_warm_start}

To isolate whether transfer gains come from the frozen background dynamics rather than from inherited perception or decoder parameters, we evaluate a perception-only warm start control, denoted as \textbf{BRICKS-WM-RBG} (BRICKS-WM with Reinitialized Background dynamics). This variant preserves the inherited encoder and decoder but reinitializes the background RSSM core before target-task training. It therefore keeps the same perceptual warm start while removing the pretrained recurrent background transition.

\begin{center}
\captionsetup{hypcap=false}
\captionof{table}{\textbf{Perception-only warm start control.} BRICKS-WM reuses the frozen background RSSM core, whereas BRICKS-WM-RBG reinitializes it while preserving inherited perception and decoding modules.}

\label{tab:perception_only_warm_start}
{\small
\setlength{\tabcolsep}{5pt}
\renewcommand{\arraystretch}{1.16}
\begin{tabular}{llccc}
\toprule
\textbf{Transfer Pair} & \textbf{Method} & \textbf{100K} & \textbf{200K} & \textbf{500K} \\
\midrule
\emph{Walker Walk} $\rightarrow$ \emph{Walker Run}
& BRICKS-WM & $247 \pm 51$ & $509 \pm 50$ & $746 \pm 4$ \\
& BRICKS-WM-RBG & $221 \pm 45$ & $472 \pm 70$ & $742 \pm 26$ \\
\midrule
\emph{Walker Walk} $\rightarrow$ \emph{Cheetah Run}
& BRICKS-WM & $465 \pm 90$ & $575 \pm 116$ & $722 \pm 78$ \\
& BRICKS-WM-RBG & $316 \pm 145$ & $543 \pm 102$ & $689 \pm 98$ \\
\midrule
\emph{Walker Walk} $\rightarrow$ \emph{Hopper Hop}
& BRICKS-WM & $0 \pm 0$ & $150 \pm 24$ & $321 \pm 100$ \\
& BRICKS-WM-RBG & $21 \pm 43$ & $98 \pm 113$ & $167 \pm 215$ \\
\midrule
\emph{Walker Run} $\rightarrow$ \emph{Walker Walk}
& BRICKS-WM & $750 \pm 131$ & $885 \pm 58$ & $940 \pm 9$ \\
& BRICKS-WM-RBG & $661 \pm 175$ & $892 \pm 34$ & $951 \pm 11$ \\
\bottomrule
\end{tabular}
}

\end{center}

Table~\ref{tab:perception_only_warm_start} shows that BRICKS-WM generally provides stronger early learning than the perception-only control. The gap is clearest in the related \emph{Walker Walk} $\rightarrow$ \emph{Walker Run} transfer and in \emph{Walker Walk} $\rightarrow$ \emph{Hopper Hop}, where reinitializing the background RSSM substantially reduces the final return. In larger-gap transfers, the reinitialized variant can partially catch up after relearning a target-specific background core. This supports the interpretation that frozen background dynamics contribute to transfer, while the benefit remains dependent on source-target compatibility.

\subsection{Early Transfer AUC}
\label{app:early_transfer_auc}

We additionally report AUC at 200K target-environment steps to quantify early learning efficiency. This metric is useful for directions such as \emph{Walker Run} $\rightarrow$ \emph{Walker Walk}, where the target task saturates quickly and final return alone can hide early transfer gains.

\begin{center}
\captionsetup{hypcap=false}
\captionof{table}{\textbf{Transfer AUC at 200K target-environment steps.} Rows denote source tasks and columns denote target tasks for transfer.}

\label{tab:transfer_auc200k}
{\small
\setlength{\tabcolsep}{6pt}
\renewcommand{\arraystretch}{1.16}
\begin{tabular}{lcccc}
\toprule
\textbf{Source $\backslash$ Target} & \textbf{Walker Walk} & \textbf{Walker Run} & \textbf{Cheetah Run} & \textbf{Hopper Hop} \\
\midrule
Walker Walk & -- & $257 \pm 14$ & $358 \pm 50$ & $29 \pm 4$ \\
Walker Run & $593 \pm 70$ & -- & $296 \pm 34$ & $0 \pm 0$ \\
Cheetah Run & $270 \pm 44$ & $104 \pm 15$ & -- & $63 \pm 51$ \\
Hopper Hop & $322 \pm 31$ & $104 \pm 7$ & $297 \pm 70$ & -- \\
\bottomrule
\end{tabular}
}

\end{center}

\begin{center}
\captionsetup{hypcap=false}
\captionof{table}{\textbf{Scratch AUC at 200K environment steps.} Scratch baselines are reported on the same target tasks.}

\label{tab:scratch_auc200k}
{\small
\setlength{\tabcolsep}{7pt}
\renewcommand{\arraystretch}{1.16}
\begin{tabular}{lcccc}
\toprule
\textbf{Method} & \textbf{Walk} & \textbf{Run} & \textbf{Cheetah} & \textbf{Hopper} \\
\midrule
DreamerV3 & $547 \pm 108$ & $245 \pm 35$ & $461 \pm 29$ & $37 \pm 39$ \\
TD-MPC2 & $477 \pm 28$ & $229 \pm 11$ & $245 \pm 41$ & $48 \pm 17$ \\
\bottomrule
\end{tabular}
}

\end{center}

As shown in Tables~\ref{tab:transfer_auc200k} and~\ref{tab:scratch_auc200k}, \emph{Walker Run} $\rightarrow$ \emph{Walker Walk} reaches $593 \pm 70$ AUC at 200K steps, exceeding both scratch baselines on \emph{Walker Walk}. The final-return curve therefore appears less visually dramatic mainly because \emph{Walker Walk} has limited remaining headroom after early saturation.

\clearpage

\begin{figure*}[t!]
\centerline{\includegraphics[width=\linewidth]{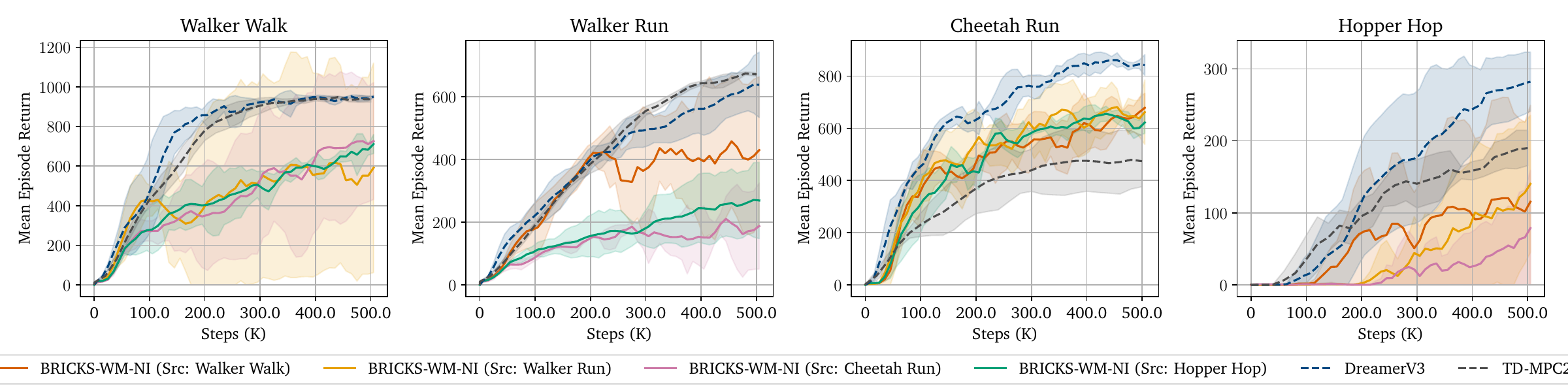}}
\caption{\textbf{Transfer performance without Interface Regularization (BRICKS-WM-NI).}}
\label{fig:no_interface_transfer}
\end{figure*}

\begin{figure*}[t!]
\centerline{\includegraphics[width=\linewidth]{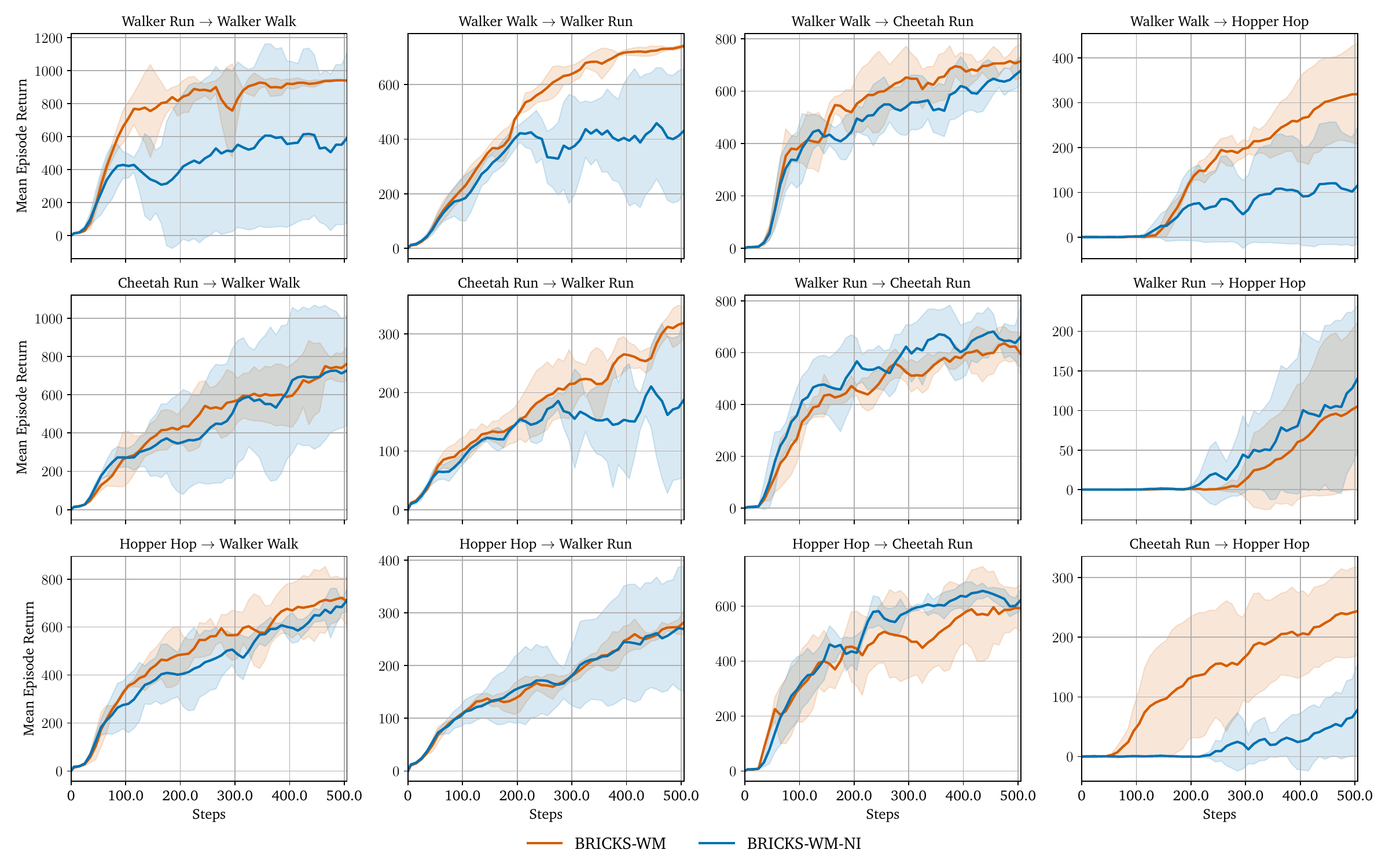}}
\caption{\textbf{Head-to-head results for interface regularization ablation on modular reuse.} }
\label{fig:interface_comparison}
\end{figure*}

\subsection{Interface Role for Transfer}
\label{app:interface_role}

To validate that interface regularization $\mathcal{R}_{\mathcal{I}}$ is essential for establishing a reusable communication protocol, we evaluate a variant of our model, denoted as \textbf{BRICKS-WM-NI} (BRICKS-WM with No Interface regularization). In this variant, we set $\beta_{\mathcal{I}}=0$ during both the source task training and the target task training, removing the information bottleneck constraint that forces the interface to filter out agent-specific details. Figure~\ref{fig:no_interface_transfer} and Figure~\ref{fig:interface_comparison} illustrate the transfer performance without interface regularization (BRICKS-WM-NI). The results indicate that without the interface regularization, the background dynamics become coupled with the specific dynamics of the source agent, leading to poor generalization on new agents. In contrast, incorporating interface regularization (BRICKS-WM) usually yields superior results compared to the unregularized ablation, demonstrating significantly better sample efficiency and final asymptotic performance. These findings empirically demonstrate that ensuring a matching interface distribution via regularization is valuable for the successful reuse of latent dynamics modules.

\subsection{Sparsity Mask Ablation}

\begin{figure*}[t!]
\centerline{\includegraphics[width=\linewidth]{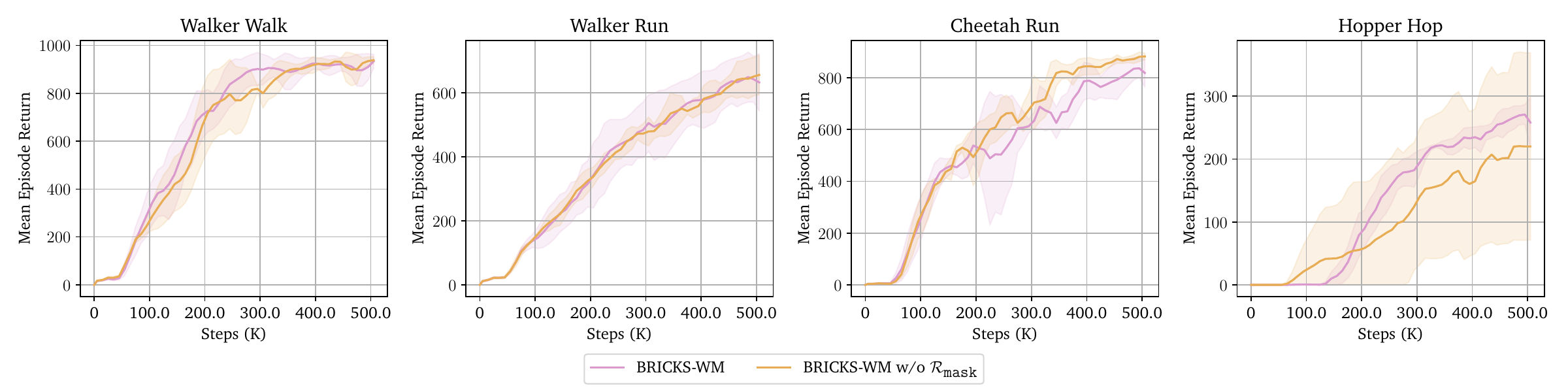}}

\caption{\textbf{Performance trained from scratch for mask regularization ablation.}}
\label{fig:mask_ablation}

\end{figure*}

In this section, we analyze the impact of the mask sparsity regularization, $\mathcal{R}_{mask}$.
Standard Slot Attention is inherently permutation equivariant, meaning the mechanism has no intrinsic preference for assigning specific entities to specific slots.
However, BRICKS-WM requires strict semantic binding, where one slot consistently represents the \textit{Agent} and the other the \textit{Background}, to enable the modular reuse of dynamics.
As shown in Figure~\ref{fig:mask_ablation}, BRICKS-WM w/o $\mathcal{R}_\texttt{mask}$ denotes training without this regularizer. The inclusion of mask regularization has a negligible impact on the performance when training from scratch. This is consistent with the intended role of $\mathcal{R}_{mask}$: it is not primarily designed as a performance-boosting regularizer, but as an identifiability-consistency bias that stabilizes the semantic binding between the Agent slot and the Background slot. The sparsity regularization serves as a necessary inductive bias to break this symmetry.
By penalizing the size of the mask generated by the designated agent slot, we leverage the bias that the agent typically occupies a smaller visual region than the background.
This constraint forces the agent slot to capture the localized foreground entity, thereby ensuring the background slot correctly captures the global environment and maintaining the functional disentanglement required for control.


\end{document}